\documentclass[pdflatex,sn-mathphys-num]{sn-jnl}% Math and Physical Sciences Numbered Reference Style
\usepackage{graphicx}%
\usepackage{multirow}%
\usepackage{amsmath,amssymb,amsfonts}%
\usepackage{amsthm}%
\usepackage{mathrsfs}%
\usepackage[title]{appendix}%
\usepackage{xcolor}%
\usepackage{textcomp}%
\usepackage{manyfoot}%
\usepackage{booktabs}%
\usepackage{caption}
\usepackage{algorithm}%
\usepackage{algorithmicx}%
\usepackage{algpseudocode}%
\usepackage{listings}%

\usepackage{footmisc}
\usepackage{graphicx}
\usepackage{amsmath}
\usepackage{bm}
\usepackage{xcolor}
\usepackage{pifont}
\usepackage{adjustbox} % for max width boxes
\usepackage{booktabs}
\usepackage{threeparttable}
\usepackage{multirow}
\usepackage{soul}
\usepackage{array}  % for \newcolumntype
\newcolumntype{P}[1]{>{\centering\arraybackslash}p{#1}}
\usepackage{algorithm}
\usepackage{algpseudocode}
\usepackage[T1]{fontenc}
\usepackage[utf8]{inputenc}

\definecolor{lightgray}{gray}{0.8} % 0=black, 1=white

\newcommand{\method}{{\fontfamily{ppl}\selectfont
R-Super}}

\theoremstyle{thmstyleone}%
\theoremstyle{thmstyletwo}%

\theoremstyle{thmstylethree}%

\begin{document}

\title[Article Title]{Scaling Artificial Intelligence for Multi-Tumor Early Detection with More Reports, Fewer Masks}


\author[1,2,3]{\fnm{Pedro R. A. S.} \sur{Bassi}}

\author[1]{\fnm{Xinze} \sur{Zhou}}
\equalcont{These authors contributed equally to this work.}

\author[1]{\fnm{Wenxuan} \sur{Li}}
\equalcont{These authors contributed equally to this work.}

\author[4]{\fnm{Szymon} \sur{Płotka}}
\equalcont{These authors contributed equally to this work.}

\author[1]{\fnm{Jieneng} \sur{Chen}}

\author[1]{\fnm{Qi} \sur{Chen}}

\author[5,14]{\fnm{Zheren} \sur{Zhu}}

\author[6]{\fnm{Jakub} \sur{Prządo}}

\author[7,8]{\fnm{Ibrahim E.} \sur{Hamaci}}

\author[9]{\fnm{Sezgin} \sur{Er}}

\author[10]{\fnm{Yuhan} \sur{Wang}}

\author[11]{\fnm{Ashwin} \sur{Kumar}}

\author[9]{\fnm{Bjoern} \sur{Menze}}

\author[12]{\fnm{Jarosław B.} \sur{Ćwikła}}

\author[10]{\fnm{Yuyin} \sur{Zhou}}

\author[11]{\fnm{Akshay S.} \sur{Chaudhari}} 

\author[11]{\fnm{Curtis P.} \sur{Langlotz}} 

\author[3]{\fnm{Sergio} \sur{Decherchi}}

\author[2,3,13]{\fnm{Andrea} \sur{Cavalli}}

\author[14]{\fnm{Kang} \sur{Wang}}

\author[14]{\fnm{Yang} \sur{Yang}}

\author[1]{\fnm{Alan L.} \sur{Yuille}}

\author*[1]{\fnm{Zongwei} \sur{Zhou}}\email{zzhou82@jh.edu}

\affil[1]{\orgname{Johns Hopkins University, Baltimore, MD, USA}}
\affil[2]{\orgname{University of Bologna, Bologna, Italy}}
\affil[3]{\orgname{Istituto Italiano di Tecnologia, Genova, Italy}}
\affil[4]{\orgname{Jagiellonian University, Kraków, Poland}}
\affil[5]{\orgname{University of California, Berkeley, CA, USA}}
\affil[6]{\orgname{Warmian-Masurian Cancer Center, Olsztyn, Poland}}
\affil[7]{\orgname{University of Zurich, Zurich, Switzerland}}
\affil[8]{\orgname{ETH AI Center, Zurich, Switzerland}}
\affil[9]{\orgname{Istanbul Medipol University, Istanbul, Turkey}}
\affil[10]{\orgname{University of California, Santa Cruz, CA, USA}}
\affil[11]{\orgname{Stanford University, Stanford, CA, USA}}
\affil[12]{\orgname{University of Warmia and Mazury, Olsztyn, Poland}}
\affil[13]{\orgname{École Polytechnique Fédérale de Lausanne, Lausanne, Switzerland}}
\affil[14]{\orgname{University of California, San Francisco, CA, USA}}

\abstract{
Early tumor detection save lives. Each year, more than 300 million computed tomography (CT) scans are performed worldwide, offering a vast opportunity for effective cancer screening. However, detecting small or early-stage tumors on these CT scans remains challenging, even for experts. Artificial intelligence (AI) models can assist by highlighting suspicious regions, but training such models typically requires extensive \textit{tumor masks}---detailed, voxel-wise outlines of tumors manually drawn by radiologists. Drawing these masks is costly, requiring years of effort and millions of dollars. In contrast, nearly every CT scan in clinical practice is already accompanied by \textit{medical reports} describing the tumor's size, number, appearance, and sometimes, pathology results---information that is rich, abundant, and often underutilized for AI training. We introduce \method, which trains AI to segment tumors that match their descriptions in medical reports. This approach scales AI training with large collections of readily available medical reports, substantially reducing the need for manually drawn tumor masks. When trained on 101,654 reports, AI models achieved performance comparable to those trained on 723 masks. Combining reports and masks further improved sensitivity by +13\% and specificity by +8\%, surpassing radiologists in detecting five of the seven tumor types. Notably, \method\ enabled segmentation of tumors in the spleen, gallbladder, prostate, bladder, uterus, and esophagus, for which no public masks or AI models previously existed. This study challenges the long-held belief that large-scale, labor-intensive tumor mask creation is indispensable, establishing a scalable and accessible path toward early detection across diverse tumor types.
We plan to release our trained models, code, and dataset at \href{https://github.com/MrGiovanni/R-Super}{https://github.com/MrGiovanni/R-Super}.
}

\maketitle

\section{Main}\label{sec1}

Cancer is a leading cause of death worldwide \cite{mathers2009global,sung2021global}. Early detection is crucial. Five-year survival rates often exceed 90\% when tumors are detected at an early stage but can drop below 20\% once the disease becomes advanced or metastatic \cite{crosby2022early}. There is no effective, widely adopted screening for these diseases, even among high-risk populations. 

Computed tomography (CT) is already part of routine care, with more than 300 million CT scans performed globally each year and 85 million in the United States alone \cite{mccollough2015answers}. These scans represent a vast, untapped opportunity for detecting tumors sooner. However, detecting tumors at an early stage from CT scans is extremely difficult, and even experienced radiologists can miss them. For instance, in a study of CT scans taken before pancreatic cancer diagnosis, about 50\% of the tumors were present but overlooked by radiologists \cite{hoogenboom2022prevalence}. 

Artificial Intelligence (AI) has the potential to help radiologists detect early tumors \cite{xia2022felix,cao2023large,hu2025ai}. AI offers several advantages: AI does not get tired or suffer from attentional effects; AI can see CT scans in 3D, while radiologists analyze them slice by slice; AI can train on large datasets (e.g., 101,654 scans in this study), surpassing the number of CT scans that a radiologist analyzes annually (est. 5,000 scans \cite{markotic2021radiologist}); and AI can see disease signs usually invisible to radiologists, such as signs of pancreatic tumors on non-contrast CT \cite{cao2023large}.
State-of-the-art AI models in tumor detection are typically formulated as \textit{semantic segmentation} \cite{bassi2025radgpt,liu2024universal,chen2023cancerunit}---a type of AI that localizes tumors and outlines them on the CT scan, accurately indicating tumor locations and boundaries, allowing radiologists to easily verify the AI's findings.

A major challenge in developing these segmentation models is the need for \textit{tumor masks}---precise outlines drawn by radiologists. Creating accurate masks is labor-intensive, costly, and not part of standard clinical workflow. Drawing a mask for each tumor can take up to 30 minutes, and one study required 8 radiologists, five years, and millions of dollars to produce 3,125 pancreatic tumor masks \cite{xia2022felix}. Public CT datasets contain tumor masks only for a few organs, such as the kidney, liver, and pancreas, with a very small number of annotated tumor scans \cite{antonelli2022medical,bilic2023liver,heller2020state,li2025pants}. For many clinically important organs, such as the spleen, gallbladder, prostate, bladder, uterus, and esophagus, no public tumor masks exist, creating a significant barrier to developing multi-tumor segmentation models.

\begin{figure}[!t]
    \centering
    \includegraphics[width=1\linewidth]{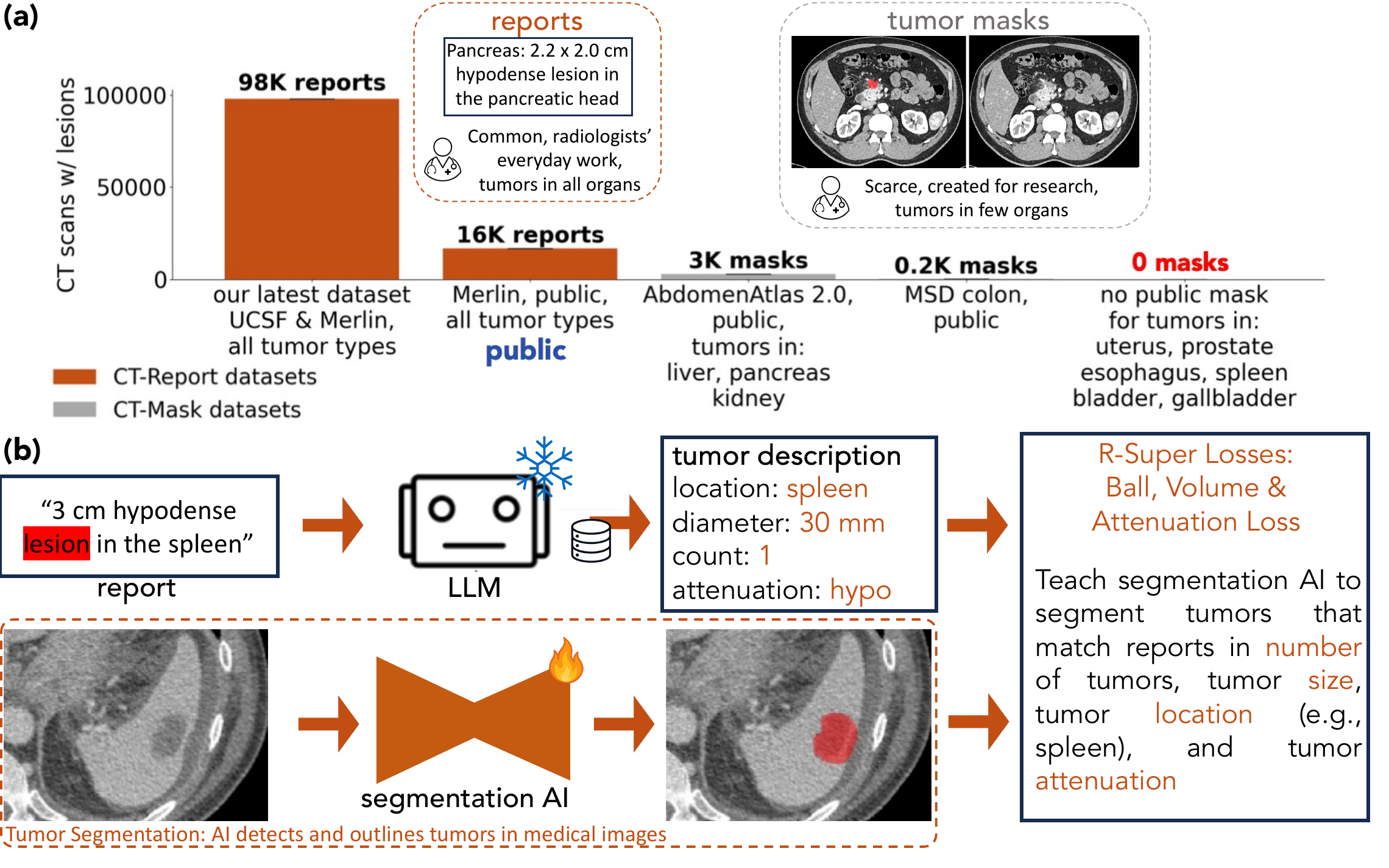}
    \caption{\textbf{(a) CT-Report datasets are much larger than CT-Mask datasets.} Our dataset has 117K CT-Report pairs, 98K with tumors. Merlin (public) \cite{blankemeier2024merlin} has 25K CT-Report pairs, 16K with tumors. In contrast, the largest CT-Mask datasets have 3K CT-Mask pairs with tumors. No tumor mask is available for many tumor types in CT. The figure also shows an example CT scan with a pancreatic tumor (PDAC), part of its report, and its tumor mask (red).
    \textbf{(b) Overview of \method\ training method.} \method\ transforms reports into per-voxel supervision for tumor segmentation, through new loss functions. It can train on both CT-Mask pairs and CT-Report pairs. For CT-Mask, \method\ uses usual dice and cross-entropy segmentation losses. For CT-Report, \method\ uses the new \textit{Volume Loss (Section \ref{sec:volume_loss}), Ball Loss (Section \ref{sec:ball_loss}) and Attenuation Loss (Section \ref{sec:att_loss})}. They optimize the segmentation output of the AI, enforcing consistency between segmented tumors and the tumor characteristics in the report---tumor count, diameters, locations, estimated volumes and attenuation. This information is extracted from the reports by an LLM and stored before training. \method\ is applicable to any segmentation architecture with minimal extra computational cost (the zero-shot LLM runs once, before training). The figure shows a spleen tumor in a CT and its segmentation by \method\ (red). %\textbf{(c) Trained on 101K CT-Report pairs, \method\ segments seven understudied tumor types.} For adrenal tumors, 53 public tumor masks exist \cite{moawad2023voxel}. For the other six tumor types, no mask exists. By learning from reports, \method\ can segment these tumors---becoming the first public AI that segments them in CT. Training with 101K CT-Report pairs matched training with 723 masks. Training with both the 101K CT-Report pairs and the 723 CT-Mask pairs provided $+11\%/+2\%/+7\%$  sensitivity/ specificity/F1-Score improvement over standard segmentation (no report). Here, we train on UCSF and Merlin, and test on UCSF test.
    }
    \label{fig:r_super}
\end{figure}

Unlike drawing tumor masks, radiologists write medical reports as part of their standard clinical workflow. These reports describe tumor  characteristics  observed in the CT scans, including the number, approximate size, location within organs, and attenuation (whether the tumor appears bright or dark), and sometimes include pathology results from biopsy or surgery. As a result, paired \textit{CT–Report} datasets are naturally much larger than \textit{CT–Mask} datasets (\figureautorefname~\ref{fig:r_super}). Public datasets \cite{hamamci2024ct2rep,blankemeier2024merlin} already provide around 25,000 CT-Report pairs, and a single hospital can easily accumulate over 500,000 CT-Report pairs (Section~\ref{sec:methods_dataset}). In contrast, public datasets rarely exceed 1,000 CT–Mask pairs \cite{antonelli2022medical,bilic2023liver,heller2020state,li2025pants}. This striking difference raises an important question: \textit{Can medical reports supplement---or even replace---tumor masks in training AI for tumor segmentation?}

Recent advances in vision–language models (VLMs) have shown capability in generating descriptive captions \cite{blankemeier2024merlin,hamamci2024ct2rep,sellergren2025medgemma,hamamci2024foundation,achiam2023gpt}. For example, models like Google’s MedGemma \cite{sellergren2025medgemma} and Stanford’s Merlin \cite{blankemeier2024merlin} can generate medical reports from CT scans. However, these models are not designed for segmentation, which requires precise tumor localization and boundary delineation. As a result, they frequently produce errors such as missing existing tumors, detecting non-existent ones, or failing to describe small and subtle lesions accurately---the very cases that are most clinically important \cite{bassi2025radgpt,chen2025vision}. A major limitation lies in their training paradigm: current VLMs rely on contrastive language–image pre-training (CLIP) \cite{radford2021learning}, which were designed to learn from generic image–text pairs like social media captions, not for the rich, structured information in radiology reports. Our approach addresses this limitation by explicitly modeling the tumor’s location as a hidden variable---similar in spirit to the Expectation–Maximization (EM) framework \cite{moon1996expectation}. The precise and descriptive nature of medical reports thus becomes a powerful supervisory signal for training. 

In this paper, we introduce \method\ (Report Supervision), which trains AI to segment tumors using radiology reports, and pathology reports when available (\figureautorefname~\ref{fig:r_super}). We then examine how much these reports can reduce the need for manual tumor masks. \method enables tumor segmentation not only in organs with many available masks but also in those with few or no tumor masks. Using \method, we trained the \textit{first} open AI model capable of segmenting tumors across seven organs\footnote{These include six tumor types with no public tumor mask in CT: spleen, gallbladder, prostate, bladder, uterus, and esophagus. We also include adrenal tumors, which have only 53 public tumor masks \cite{moawad2023voxel}. No public AI model can segment these seven tumor types.}. The key innovation of \method\ lies in report-supervised loss functions that directly teach the AI to segment tumors consistent with the tumor descriptions in reports—in terms of tumor count, size, location\footnote{A tumor location in a report is provided as the organ, organ sub-segment, and/or slice where the tumor is. A slice is a plane localizing the tumor in the 3D CT scan.}, and attenuation. 
Conceptually, this involves learning from incomplete data: reports describe many tumor characteristics but not the exact tumor outline, so \method\ teaches the segmentation model to estimate outlines consistent with the available tumor  characteristics  in reports.
By using reports to guide tumor segmentation, unlike prior approaches (e.g., VLMs), \method\ learns efficiently and achieves superior tumor detection performance (Table~\ref{tab:multi_tumor_results}). 
\method\ extracts the tumor characteristics from reports using large language models (LLMs) with radiologist-designed prompts and store it before training. Importantly, reports are only used in training, not in inference. 
\method\ can train any segmentation model architecture, and it can learn from just CT-Report pairs. To further improve accuracy, \method\ can also learn from CT-Mask pairs together with the CT-Report pairs. Thus, \method\ can segment tumors without public masks, and it can further scale the largest CT-Mask datasets (e.g., PanTS \cite{li2025pants}) with many CT-Report pairs.

To train \method\ to segment multiple tumor types lacking public tumor masks\footnote{Our dataset includes benign, primary (malignant) and metastatic tumors.}, we created \textit{the largest CT-Report training dataset to date}---101,654 CT-Report pairs (Section \ref{sec:methods_dataset} and \tableautorefname~\ref{tab:datasets_summary}). These CT scans were performed in the University of California San Francisco (UCSF) hospital and affiliated institutions during the last 28 years. Our dataset also includes the public Merlin dataset \cite{blankemeier2024merlin} (25,494 CTs, Stanford Hospital, from 2012 to 2018). To the best of our knowledge, no previous study used 100,000+ CT-Report pairs to train AI. First, we used these 101,654 CT-Report pairs to train \method. Then, to further improve performance, we created tumor masks for our dataset, and trained \method\ on both the CT-Report and CT-Mask pairs. To create these tumor masks efficiently, we introduced a report-guided active learning cycle (Section \ref{sec:methods_dataset}): (I) \method\ automatically created tumor masks for our dataset; (II) we identified the most incorrect tumor masks by comparing them to reports; (III) these incorrect tumor masks were revised by 31 radiologists; (IV) we trained \method\ using the revised tumor masks and all CT-Report pairs. We repeated this cycle until reaching 723 radiologist-corrected tumor masks. The radiologists reported that our report-guided active learning cycle reduced the average time to create each mask from about 30 to five minutes. Learning jointly from CT–Report and CT–Mask pairs, \method\ achieves substantially higher performance in comparison to standard segmentation training with just CT-Mask pairs---both when few masks are available (first active learning cycles), or when many are available (final cycles).

We evaluated \method\ through internal and external validation on three datasets. Internal validation used unseen patients from the hospitals in our training dataset, while external validation tested \method\ in a hospital never seen during training. \method\ accurately detected tumors in seven organs lacking public tumor masks. In five of these tumor types, \method\ surpassed radiologist tumor detection performances reported in the literature (\tableautorefname~\ref{tab:multi_tumor_results})\footnote{We compare our results with radiologist tumor detection performance reported in the literature. Our AI was evaluated on a dataset containing both healthy patients and those with malignant tumors, and we searched the literature for studies that also assessed radiologists on datasets with healthy and malignant tumor patients. However, this comparison remains limited, as the AI and radiologists were tested on different datasets—including distinct patient populations, CT scanners, and tumor characteristics (see Appendix \ref{app:radiologist_studies} for an analysis of the selected studies and limitations of our comparison between radiologists and AI). Consequently, these comparisons offer a qualitative sense of the detection difficulty across tumor types. A more rigorous comparison would require a reader study in which radiologists and AI are assessed on the same dataset.} In tumor detection, \method\ consistently outperformed VLMs such as Merlin \cite{blankemeier2024merlin} (Stanford University) and MedGemma \cite{sellergren2025medgemma} (Google) by double digit margins  (\tableautorefname~\ref{tab:multi_tumor_results}). %\ay{It outperformed VLMs for what task -- detection? -- clarify because ours can do location/segmentation as well.} 
Trained with 101,654 CT–Report pairs and 723 CT–Mask pairs, \method\ exceeded standard segmentation (trained only on the 723 CT–Mask pairs our radiologists created) by margins of +13/+8\% in sensitivity/specificity (\figureautorefname~\ref{fig:merlin_test}). Importantly, these outperformance
%\ay{outperformance?} 
margins were also large for detecting small tumors (< 2 cm): +7/+5.3\%. \method\ improved performance when both few (e.g., 52, Section \ref{sec:multi_tumor_results}) and many tumor masks (e.g., 900, Section \ref{sec:pants}) were available for training. 
Remarkably, when trained only with CT–Report pairs (620 to 10,980 per tumor type), \method\ surpassed segmentation trained with few masks (52 to 185 per tumor type, Section \ref{sec:external_eval}). This shows that large-scale weak supervision (reports) can outperform small-scale strong supervision (tumor masks) in tumor segmentation, echoing strong trends in computer vision \cite{radford2015unsupervised} and natural language processing \cite{achiam2023gpt}. Our main contributions are:

%\TODO{stress more that our performance is reproducible? private studies are not reproducible. Stress the problem in private studies: if you just create masks and get a performance but nothing is made public, no benefit to community---in the discussion, you may cite papers that have private models, you cannot test in your dataset, not reproducible. Ask Alan and Zongwei, should we cite private results?}

\begin{enumerate}
\item \textbf{\method:} a new AI training method that enforces consistency between tumors segmented by AI and report descriptions of these tumors (tumor number, size, location, and attenuation). It can train any segmentation architecture using CT–Report pairs alone or CT–Report \& CT–Mask pairs. \method\ has the first loss functions that directly supervise CT tumor segmentation using reports (\figureautorefname~\ref{fig:r_super}).

\item \textbf{Early tumor detection:} by learning from 101,654 readily available radiology reports, \method\ improves the detection of small tumors by double-digit margins (\tableautorefname~\ref{tab:small_tumors}), showing potential to enhance early cancer detection—critical for survival.

\item \textbf{Enabling multi-tumor segmentation and open science:} we release the first public segmentation model capable of segmenting seven tumor types lacking public segmentation masks in CT. It surpasses reported radiologist performance in four tumor types (\tableautorefname~\ref{tab:multi_tumor_results}). We also release CT, tumor masks and reports for these tumors, giving the community methods and data to segment understudied tumor types and advance opportunistic, multi-organ tumor detection in real-world CT scans.
\end{enumerate}

This paper builds on our prior conference paper \cite{bassi2025learning}, providing several improvements: (1) the \method\ loss functions now also use the tumor slice and attenuation information from reports---exploiting all tumor characteristics in most reports; (2) we now segment seven tumor types without public masks, previously we segmented only pancreatic and kidney tumors, which exist in public CT-Mask datasets; and (3) we scaled our training dataset from 6,718 to 101,654 CT-Report pairs, and 31 radiologists created 723 tumor masks for it. This study is about tumor segmentation, but as an addendum, we also trained our AI to produce CLIP embeddings—giving the community the first public AI trained on 100,000+ CTs to create such embeddings---used for tasks such as report generation \cite{blankemeier2024merlin}.% \ay{why should the community want CLIP embeddings? -- give a use case.}

%\ay{Good to stress the size of the datasets -- but saying how this method (mostly) outperforms published reported results for radiologists will probably make it more interesting for a Nature reader. Also to say that active learning can be used to improve the results by annotating more tumor maksks efficiently}

\section{Results}

\subsection{Validation Methodology \& Dataset Overview}

%\TODO{Mention pathology confirmation and malignant on test sets, update test set sizes in table}

We perform two kinds of validation: report-based validation and mask-based validation. In report-based validation (Section \ref{sec:multi_tumor_results}, \ref{sec:small_tumor_results}, and \ref{sec:external_eval}) we use radiology reports as ground-truth and evaluate tumor detection at the organ-level. I.e., we compare the segmentation model output and the report, checking for tumor absence/presence in each organ and calculating tumor detection sensitivity, specificity and F1-Score. E.g., if a report mentions a tumor in the spleen, the AI is correct if it segmented a spleen tumor. 
For report-based evaluation, we transform the tumor segmentation outputs generated by segmentation models (like R-Super) into categorical outputs (e.g., spleen tumor present/absent). To do so, we use voxel-count and confidence thresholds. For example, with a voxel count threshold of 50 and a confidence threshold of 50\%, we consider that the segmentation model predicted a spleen tumor if more than 50 voxels of the "spleen tumor" class have more than 50\% confidence (after sigmoid activation). In mask-based validation (Section \ref{sec:pants}) we take advantage of ground truth tumor masks to perform the regular segmentation validation, using Dice Similarity Coefficient (DSC) and Normalized Surface Distance (NSD). \tableautorefname~\ref{tab:datasets_summary} explains which datasets were used in which parts of the following sections.

\setlength{\tabcolsep}{1pt}

\begin{table}[t]
\centering
\scriptsize
\begin{tabular*}{\linewidth}{@{\extracolsep{\fill}} lcccccccc @{}}
\toprule
dataset & total & spleen & esophagus & bladder & gallbladder & adrenal & uterus & prostate \\
\midrule
UCSF \& Merlin train & 101,654 & 11,677 & 541 & 3,271 & 2,892 & 9,609 & 5,075 & 1,902 \\
UCSF & 85,899 & 10,980 & 620 & 3,167 & 2,628 & 8,996 & 4,367 & 1,948 \\
Merlin & 25,469 & 1,820 & 107 & 565 & 683 & 1,716 & 1,359 & 404 \\
UCSF test & 1,220 & 181 & 72 & 112 & 77 & 263 & 100 & 158 \\
%UCSF test small & 951 & 196 & 6 & 52 & 51 & 235 & 53 & 50 \\
Merlin test & 1,133 & 135 & 22 & 107 & 57 & 544 & 58 & 38 \\
Masks UCSF \& Merlin & 723 & 87 & 185 & 88 & 75 & 52 & 169 & 67 \\
Masks UCSF & 612 & 66 & 183 & 63 & 52 & 29 & 156 & 63 \\
%Masks Merlin test & 111 & 21 & 2 & 25 & 23 & 23 & 13 & 4 \\
\bottomrule
\end{tabular*}
\caption{\textbf{Our CT-Report training dataset (UCSF \& Merlin train) has an unprecedented size: 101.6K CT-Report pairs.} Importantly, it focuses on seven understudied tumor types. To the best of our knowledge, no previous research project trained AI on 100K+ CT-Report pairs. The table illustrates the number of CT scans with each type of tumor. Section \ref{sec:multi_tumor_results} and Section \ref{sec:small_tumor_results} used the datasets 'UCSF \& Merlin train' and 'Masks UCSF \& Merlin' for training, and 'UCSF test' for testing. Section \ref{sec:external_eval} used UCSF for training and 'Merlin test' for testing. Section \ref{sec:pants} trains on the public PanTS dataset \cite{li2025pants} (9K CT-Mask pairs, and 0.9K with pancreatic tumors) and the public Merlin dataset (we selected 1.8k CT-Report pairs with pancreatic tumors, plus 1.8K normals). In Section \ref{sec:pants}, we test on the PanTS test set (901 CT-Mask pairs, 151 with pancreatic tumors), and a Merlin test set with 400 CT-Report pairs, 200 with pancreatic tumors. Merlin and PanTS are already publicly available. The PanTS datasets and all our training datasets include normals, benign tumors and malignant tumors. The Merlin and UCSF test datasets include only malignant tumors (primary and metastasis) and healthy patients. Malignancy is confirmed by explicit mentions of malignancy in radiology reports, or by pathology reports (available in UCSF test).}
\label{tab:datasets_summary}
\end{table}
%gathering data

Report-based validation allows for large-scale test datasets, because it does not require ground-truth tumor masks. Moreover, it allows for comparisons between segmentation models and other types of AI, such as VLMs\footnote{To evaluate VLMs, we ask them to write radiology reports, and we automatically analyze whether these reports indicate tumor presence or absence in each organ, following \cite{bassi2025radgpt}.}. %It also allows comparing segmentation models to radiologists, since the literature informs sensitivity and specificity of radiologists. 
On the other hand, mask-based validation is more precise than report-based validation, using DSC and NSD to evaluate how well segmentation models outline tumors. However, it incurs smaller test datasets, because it needs ground-truth tumor masks. Also, we can only calculate DSC and NSD for segmentation models, not for VLMs. Therefore, we use both mask-based and report-based validation.

R-Super can train any segmentation architecture. We used MedFormer (a U-Net-based convolutional neural network and transformer hybrid \cite{gao2022data}) as the segmentation architecture for R-Super and the standard segmentation model. % all alternative segmentation models (multi-task learning, standard segmentation without reports, models genesis and pseudo-labels).
MedFormer was chosen because of its strong performance---top 1 position in the Touchstone Segmentation Benchmark \cite{bassi2024touchstone}. Public AI models use their original architectures: medgemma-4b-it for MedGemma, RadLlama-7B for Merlin, and nnU-Net \cite{isensee2021nnu} for ULS.

%\TODO{Dataset plot: number of reports and masks per tumor type, separate large and small, training and testing, mask and report. Maybe a table?}

\subsection{\method\ Detects seven Tumor Types w/o Public Mask}
\label{sec:multi_tumor_results}

\begin{figure}[t]
    \centering
    \includegraphics[width=1\linewidth]{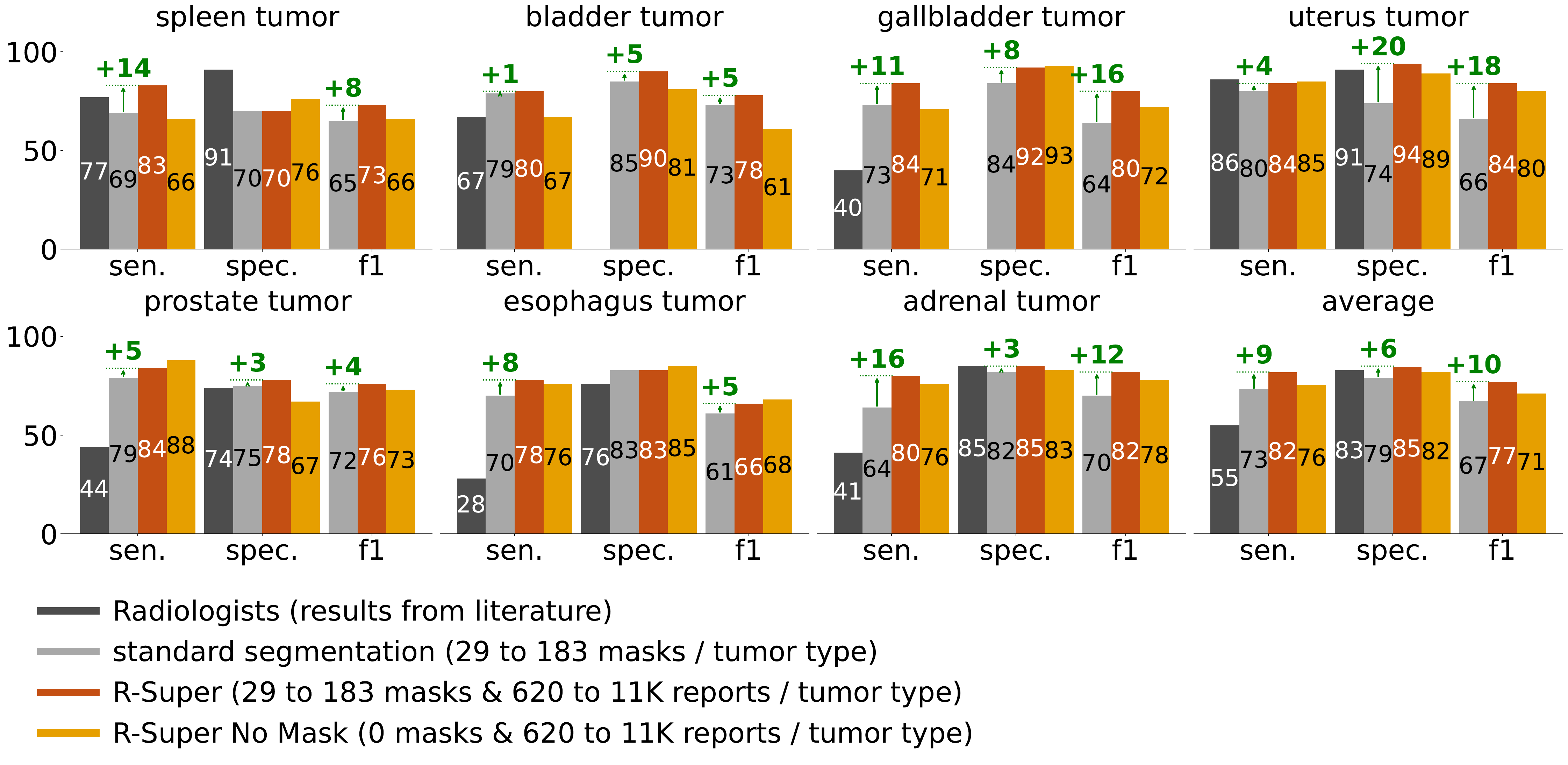}
    \caption{\textbf{Trained on 101K CT-Report pairs, \method\ segments seven understudied tumor types.} For adrenal tumors, 53 public tumor masks exist \cite{moawad2023voxel}. For the other six tumor types, no mask exists. By learning from reports, \method\ can segment these tumors---becoming the first public AI that segments them in CT. \method\ surpasses radiologist tumor detection performance for five of the seven tumor types. Radiologist performance was extracted from the literature, see Appendix \ref{app:radiologist_studies} for an analysis of the selected studies and limitations of the comparison. Training with 101K CT-Report pairs surpassed training with 723 masks, showing that large-scale weak supervision (many reports) can surpass small-scale strong supervision (few masks). Training with both the 101K CT-Report pairs and the 723 CT-Mask pairs provided $+9\%/+6\%/+10\%$  sensitivity/ specificity/F1-Score improvement over standard segmentation (no report). Here, we train on UCSF and Merlin, and test on UCSF test---N=1,220 (see dataset descriptions in \tableautorefname~\ref{tab:datasets_summary}). Radiologist performances in tumor detection were acquired from several studies: }
    \label{fig:results_ucsf}
\end{figure}

%\TODO{add radiologists in plots and captions. Make your models the first in the plots}

\setlength{\tabcolsep}{1pt}
\renewcommand{\arraystretch}{0.9}

\begin{table*}[t!]
\centering
\caption{\textbf{With \method, reports enable tumor detection across seven types of tumors unavailable in public datasets.} %We compared \method\ to %other training methods that can learn from reports---multi-task learning (MTL), report-refined pseudo labels, and contrastive CT-report pretraining (CLIP-like); to standard segmentation; and to self-supervised pretraining (Models Genesis). These methods were trained with the same dataset as R-Super: 723 CT-Mask pairs, plus 101,654 CT-Report pairs (for training methods that can learn from reports). Models Genesis was pre-trained on 101,654 CT scans and fine-tuned on CT-Mask pairs. We also compare \method\ standard segmentation (training from masks only) to publicly available AI checkpoints: 
\method\ surpasses radiologist tumor detection performance for five of the seven tumor types. Radiologist performance was extracted from the literature, see Appendix \ref{app:radiologist_studies} for an analysis of the selected studies and limitations of the comparison. Results include \method\ trained with reports (101K) and masks (723), and \method\ trained with reports only. We also compare it to standard segmentation (trained with only masks) and to public AI models: ULS, a nnU-Net \cite{isensee2021nnu} segmentation model trained for universal lesion segmentation on lesions in unspecified organs; MedGemma, the flagship medical VLM from Google; and Merlin, the latest medical VLM from Stanford University. %We include two versions of Merlin: the Merlin report generation model and Merlin with last-layer fine-tuning for tumor detection. Fine-tuning was conducted on the R-Super dataset. 
We test on 1,220 CT scans from UCSF (internal validation), using reports as ground truth (no masks available). See dataset details in \tableautorefname~\ref{tab:datasets_summary}. For DSC and NSD, see Section \ref{sec:pants}.}
\label{tab:multi_tumor_results}

%\TODO{change test dataset size in captions}

%\TODO{Correct numbers in captions (masks and reports)}
%\TODO{Remove comparison to radiologists}
%\TODO{\szymon{Maybe bold the best value?}}

\begin{adjustbox}{max width=\textwidth}
\scriptsize
\begin{tabular*}{\textwidth}{@{\extracolsep{\fill}}%
  p{0.25\textwidth}
  *{8}{P{0.045\textwidth}}
  @{}}
\toprule
& \multicolumn{2}{c}{bladder}
& \multicolumn{2}{c}{esophagus}
& \multicolumn{2}{c}{gallbladder}
& \multicolumn{2}{c}{uterus}
\\
\cmidrule(lr){2-3}\cmidrule(lr){4-5}\cmidrule(lr){6-7}\cmidrule(lr){8-9}
AI
& sen. & spe.
& sen. & spe.
& sen. & spe.
& sen. & spe.
\\
\midrule
radiologists (literature) & 67.0 & n/a & 28.0 & 76.0 & 40.0 & n/a & 86.0 & 91.0 \\
\multicolumn{9}{l}{\textit{Public Vision-Language Models}}\\
Merlin \cite{blankemeier2024merlin} & 1.9 & 99.6 & 0.0 & 100.0 & 1.4 & 98.5 & 0.0 & 100.0 \\
MedGemma \cite{sellergren2025medgemma} & 2.9 & 100.0 & 0.0 & 100.0 & 0.0 & 100.0 & 1.0 & 99.6 \\
\midrule
\multicolumn{9}{l}{\textit{Public Universal Lesion Segmentation Models}}\\
ULS \cite{de2025uls23} & 57.7 & 72.5 & 28.6 & 92.4 & 23.6 & 96.6 & 39.3 & 85.9 \\
\midrule
standard segmentation & 78.8 & 85.2 & 70.1 & 82.9 & 72.6 & 84.4 & 80.4 & 74.1 \\
\midrule
\textbf{R-Super No Mask} & 66.7 & 80.9 & 76.2 & 85.3 & 71.1 & 92.8 & 85.2 & 88.8 \\
\textbf{R-Super} & 79.5 & 89.5 & 77.8 & 83.3 & 84.4 & 91.8 & 84.0 & 94.2 \\
\bottomrule
\end{tabular*}
\end{adjustbox}

\vspace{0.6em}

\begin{adjustbox}{max width=\textwidth}
\scriptsize
\begin{tabular*}{\textwidth}{@{\extracolsep{\fill}}%
  p{0.25\textwidth}
  *{8}{P{0.045\textwidth}}
  @{}}
\toprule
& \multicolumn{2}{c}{prostate}
& \multicolumn{2}{c}{adrenal}
& \multicolumn{2}{c}{spleen}
& \multicolumn{2}{c}{average}
\\
\cmidrule(lr){2-3}\cmidrule(lr){4-5}\cmidrule(lr){6-7}\cmidrule(lr){8-9}
AI
& sen. & spe.
& sen. & spe.
& sen. & spe.
& sen. & spe.
\\
\midrule
radiologists (literature) & 44.0 & 74.0 & 41.1 & 84.5 & 76.9 & 90.9 & 54.7 & 83.3 \\
\multicolumn{9}{l}{\textit{Public Vision-Language Models}}\\
Merlin \cite{blankemeier2024merlin} & 0.6 & 99.2 & 0.0 & 99.6 & 0.6 & 98.5 & 0.6 & 99.3 \\
MedGemma \cite{sellergren2025medgemma} & 0.0 & 100.0 & 1.6 & 99.6 & 7.2 & 96.0 & 1.8 & 99.3 \\
\midrule
\multicolumn{9}{l}{\textit{Public Universal Lesion Segmentation Models}}\\
ULS \cite{de2025uls23} & 33.7 & 84.7 & 0.0 & 100.0 & 20.1 & 85.9 & 29.0 & 88.3 \\
\midrule
standard segmentation & 79.1 & 75.2 & 63.8 & 81.7 & 69.1 & 70.2 & 73.4 & 79.1 \\
\midrule
\textbf{R-Super No Mask} & 88.4 & 66.9 & 75.6 & 83.1 & 65.8 & 76.3 & 75.6 & 82.0 \\
\textbf{R-Super} & 83.5 & 77.5 & 79.5 & 85.2 & 82.9 & 69.6 & 81.7 & 84.4 \\
\bottomrule
\end{tabular*}
\end{adjustbox}

\end{table*}

%\TODO{Cite the radiologist studies in literature}

\method\ segments tumors in the spleen, gallbladder, prostate, bladder, uterus, esophagus and adrenal glands. No public tumor masks for these organs exist, except for adrenal gland tumors, which have only 53 public masks \cite{moawad2023voxel}. \tableautorefname~\ref{tab:multi_tumor_results} shows that \method, trained with 723 CT-Mask pairs \& 101,654 CT-Report pairs, surpasses public VLMs such as Google's MedGemma and Stanford's Merlin by large margins. All VLMs struggled to find tumors, generating radiology reports of low tumor detection sensitivity. VLMs did not surpass a standard segmentation model (trained with only CT-Mask pairs), as seen in previous studies \cite{bassi2025radgpt}. The VLM results here are worse than in \cite{bassi2025radgpt}, possibly for two reasons: the tumors we consider here are rarer than liver, kidney and pancreas tumors, considered in \cite{bassi2025radgpt}, and/or they are more difficult to detect. The Universal Lesion Segmentation (ULS) model also underperformed (33.7\% average tumor detection F1-Score). This likely reflects limitations in training data: the ULS dataset does not distinguish tumor types, and the rare tumors considered here were possibly underrepresented, hindering accurate segmentation. %Furthermore, it substantially surpasses other AI training methods that can learn from reports---multi-task learning, CLIP, Models Genesis, pseudo-labels---even though they were trained in the same dataset as \method. 
\method\  has the best performance in \tableautorefname~\ref{tab:multi_tumor_results}, surpassing the standard segmentation model by large margins: +9\%/+6\%/+10\% in sensitivity/specificity/F1-Score (\figureautorefname~\ref{fig:results_ucsf}). Therefore, unlike VLMs, R-Super effectively used reports to improve tumor detection.
%\TODO{add numbers in this text}
%\TODO{Our dataset number are not adding up}
%\TODO{statistical tests}
%\TODO{comparison to humans}
%\TODO{check results in text, see if all is updated.}
%why no dice
%add numbers
%compare to radiologists

Even when trained with only CT-Report pairs (no masks), \method\ surpassed standard segmentation trained with only CT-Mask pairs (no reports). Therefore, many reports (541 to 11.7K per tumor type, \tableautorefname~\ref{tab:datasets_summary}) can offer more training value than few masks (52 to 185 per tumor type, \tableautorefname~\ref{tab:datasets_summary}). This discovery may seem surprising in the field of tumor segmentation. However, in other AI fields, like Natural Language Processing (NLP) and computer vision, weaker supervision in large-scale can also surpass stronger supervision at smaller-scale \cite{zhuang2025mim,zhuang2025advancing}. In NLP, powerful LLMs like ChatGPT were only possible after the transition from a small text dataset with precise labels to massive (billion-scale) text datasets with weaker labels (self-supervision). Similarly, in computer vision, VLMs that can understand images and generalize to multiple tasks and domains were only possible after the transition from small image datasets with precise classification labels to massive image datasets with weaker labels (captions). Our results suggest that the transition from small CT-Mask datasets to massive CT-Report datasets may also transform the tumor segmentation field.

The best performance in \tableautorefname~\ref{tab:multi_tumor_results} is achieved by \method\ trained on CT-Mask pairs (723) plus CT-Report pairs (101,654). \method\ can segment tumors with zero masks, and it gets better when more masks become available. This result makes \method\ an efficient tool to accelerate mask creation with active learning---a cycle where AI creates masks, radiologists correct the worst AI-made masks, and AI retrains on the corrected masks (getting better). \method\ helps at every step: it can train on only CT-Report pairs to help radiologists produce the initial masks, and it can generate increasingly better masks as more masks become available for training, helping the radiologists more. Indeed, we used \method\ in a active learning loop to help radiologists create the 723 masks in our dataset (see Section \ref{sec:methods_dataset}).

For five of the seven tumor types in Table \ref{tab:multi_tumor_results}, \method\ surpassed the tumor detection performance of radiologists (for tumors in the bladder, gallbladder, uterus, prostate, esophagus, and adrenal glands). For these tumor types, CT scans are not the primary diagnostic tool---since these tumor types usually are difficult to see in CT. However, our results show that AI may be able too see tumors signs that are not easily perceptible to humans. This echoes with previous studies, which show that AI can see pancreatic tumors in non-contrast CT with high-accuracy, but humans cannot \cite{reiss2021panda}. This finding is especially relevant for opportunistic detection: with >300 million CT scans performed annually for diverse clinical reasons, segmentation models have the potential to scan images in the background, flag suspicious studies and regions, and prompt radiologists to review those areas and refer patients for targeted follow-up when needed.

We drew radiologist performances from published studies that—where possible—tested tumor detection on datasets containing both healthy and malignant tumor patients, such as our test dataset. Some caveats limit head-to-head comparisons: the bladder \cite{malik2023systematic} and gallbladder \cite{frezza1997gallbladder} studies lacked normal controls (so only sensitivity is available); the esophagus study \cite{sui2021detection} considered non-contrast CT, but our test set uses contrast-enhanced CT (where tumors are easier to detect); the bladder study \cite{malik2023systematic} considered pre-diagnostic CT, where tumor detection is more difficult; and the adrenal gland study \cite{allard1990sensitivity} considered only metastatic adrenal tumors, while our AI has both primary and metastatic adrenal tumors. Descriptions of all selected studies and comparison limitations appear in Appendix \ref{app:radiologist_studies}.

\subsection{\method\ Detects Small Tumors}
\label{sec:small_tumor_results}

\begin{table*}[t]
\centering
\caption{\textbf{With \method, reports improve small tumor detection across seven types of tumors unavailable in public datasets.} The detection of small tumors is crucial because it can improve early cancer detection and patient survival. 
%We compare \method\ to other AI training methods that can learn from reports, to standard segmentation; and to self-supervised pretraining (Models Genesis). These methods were trained with the same dataset as R-Super: 723 CT-Mask pairs, plus 101,654 CT-Report pairs (for training methods that can learn from reports). Models Genesis was pre-trained on 101,654 CT scans and fine-tuned on CT-Mask pairs. We also compare \method\ to publicly available AI checkpoints: ULS, trained for universal lesion detection on lesions in unspecified organs, MedGemma, the flagship medical AI from Google (VLM), and Merlin, the latest medical VLM from Stanford University. 
Results include \method\ trained with reports (101K) and masks (723), and \method\ trained with reports only. We compare it to standard segmentation (trained with only masks) and to public AI models: ULS, a nnU-Net \cite{isensee2021nnu} segmentation model trained for universal lesion segmentation on lesions in unspecified organs; MedGemma, the flagship medical VLM from Google; and Merlin, the latest medical VLM from Stanford University. %We include two versions of Merlin: the Merlin report generation model, and Merlin with last-layer fine-tuning for tumor detection. Fine-tuning was conducted on the R-Super dataset. 
We test on 470 CT scans from UCSF (internal validation), 257 healthy and 213 with small tumors ($<$ 2 cm diameter). We use pathology reports as ground truth (no masks available). See data details in \tableautorefname~\ref{tab:datasets_summary}. For DSC and NSD, see Section \ref{sec:pants}.}
\label{tab:small_tumors}

\begin{adjustbox}{max width=\textwidth}
\scriptsize
\begin{tabular*}{\textwidth}{@{\extracolsep{\fill}}%
  p{0.25\textwidth}
  *{8}{P{0.045\textwidth}}
  @{}}
\toprule
& \multicolumn{2}{c}{bladder}
& \multicolumn{2}{c}{esophagus}
& \multicolumn{2}{c}{gallbladder}
& \multicolumn{2}{c}{uterus}
\\
\cmidrule(lr){2-3}\cmidrule(lr){4-5}\cmidrule(lr){6-7}\cmidrule(lr){8-9}
AI
& sen. & spe.
& sen. & spe.
& sen. & spe.
& sen. & spe.
\\
\midrule
\multicolumn{9}{l}{\textit{Public Vision-Language Models}}\\
Merlin \cite{blankemeier2024merlin} & 0.0 & 99.6 & 0.0 & 100.0 & 0.0 & 98.5 & 0.0 & 100.0 \\
MedGemma \cite{sellergren2025medgemma} & 7.1 & 100.0 & 0.0 & 100.0 & 0.0 & 100.0 & 0.0 & 99.6 \\
\midrule
\multicolumn{9}{l}{\textit{Public Universal Lesion Segmentation Models}}\\
ULS \cite{de2025uls23} & 66.7 & 72.5 & 0.0 & 92.4 & 33.3 & 96.6 & 33.3 & 85.9 \\
\midrule
standard segmentation & 46.7 & 85.2 & 80.0 & 82.9 & 28.6 & 84.4 & 100.0 & 74.1 \\
\midrule
\textbf{R-Super No Mask} & 26.7 & 80.9 & 100.0 & 85.3 & 42.9 & 92.8 & 66.7 & 88.8 \\
\textbf{R-Super} & 68.8 & 89.5 & 80.0 & 83.3 & 37.5 & 91.8 & 66.7 & 94.2 \\
\bottomrule
\end{tabular*}
\end{adjustbox}

\vspace{0.6em}

\begin{adjustbox}{max width=\textwidth}
\scriptsize
\begin{tabular*}{\textwidth}{@{\extracolsep{\fill}}%
  p{0.25\textwidth}
  *{8}{P{0.045\textwidth}}
  @{}}
\toprule
& \multicolumn{2}{c}{prostate}
& \multicolumn{2}{c}{adrenal}
& \multicolumn{2}{c}{spleen}
& \multicolumn{2}{c}{average}
\\
\cmidrule(lr){2-3}\cmidrule(lr){4-5}\cmidrule(lr){6-7}\cmidrule(lr){8-9}
AI
& sen. & spe.
& sen. & spe.
& sen. & spe.
& sen. & spe.
\\
\midrule
\multicolumn{9}{l}{\textit{Public Vision-Language Models}}\\
Merlin \cite{blankemeier2024merlin} & 7.1 & 99.2 & 0.0 & 99.6 & 0.0 & 98.5 & 1.0 & 99.3 \\
MedGemma \cite{sellergren2025medgemma} & 0.0 & 100.0 & 2.0 & 99.6 & 7.4 & 96.0 & 2.4 & 99.3 \\
\midrule
\multicolumn{9}{l}{\textit{Public Universal Lesion Segmentation Models}}\\
ULS \cite{de2025uls23} & 21.4 & 84.7 & 0.0 & 100.0 & 6.8 & 85.9 & 23.1 & 88.3 \\
\midrule
standard segmentation & 64.3 & 75.2 & 60.7 & 81.7 & 54.2 & 70.2 & 62.1 & 79.1 \\
\midrule
\textbf{R-Super No Mask} & 64.3 & 66.9 & 70.6 & 83.1 & 49.2 & 76.3 & 60.1 & 82.0 \\
\textbf{R-Super} & 71.4 & 77.5 & 79.4 & 85.2 & 80.0 & 69.6 & 69.1 & 84.4 \\
\bottomrule
\end{tabular*}
\end{adjustbox}

\end{table*}

%\TODO{mention radiologist performance in text too}

\method\ also surpassed the state-of-the-art in the detection
of tumors smaller than 2 cm in diameter (see \tableautorefname~\ref{tab:small_tumors}). The detection of small tumors is especially important for early cancer detection and better patient survival. 
However, it is very challenging, as small tumors occupy as little as 0.0001\% of a CT scan volume \cite{ardila2019end,mikhael2023sybil,cao2025boosting}.
To address this challenge, the \method\ loss functions transform reports into per-voxel supervision concentrated on the organ where the tumor is, or even on a small part of this organ (Section \ref{sec:ball_loss}).
This strategy was successful: in comparison to standard segmentation (trained with CT-Mask pairs, no report), \method\ (trained with CT-Report and CT-Mask pairs) yielded an improvement of +7\%/+5.3\% in tumor detection sensitivity/specificity.

\subsection{\method\ Generalizes to Unseen Hospitals}
\label{sec:external_eval}

\begin{figure}[t]
    \centering
    \includegraphics[width=1\linewidth]{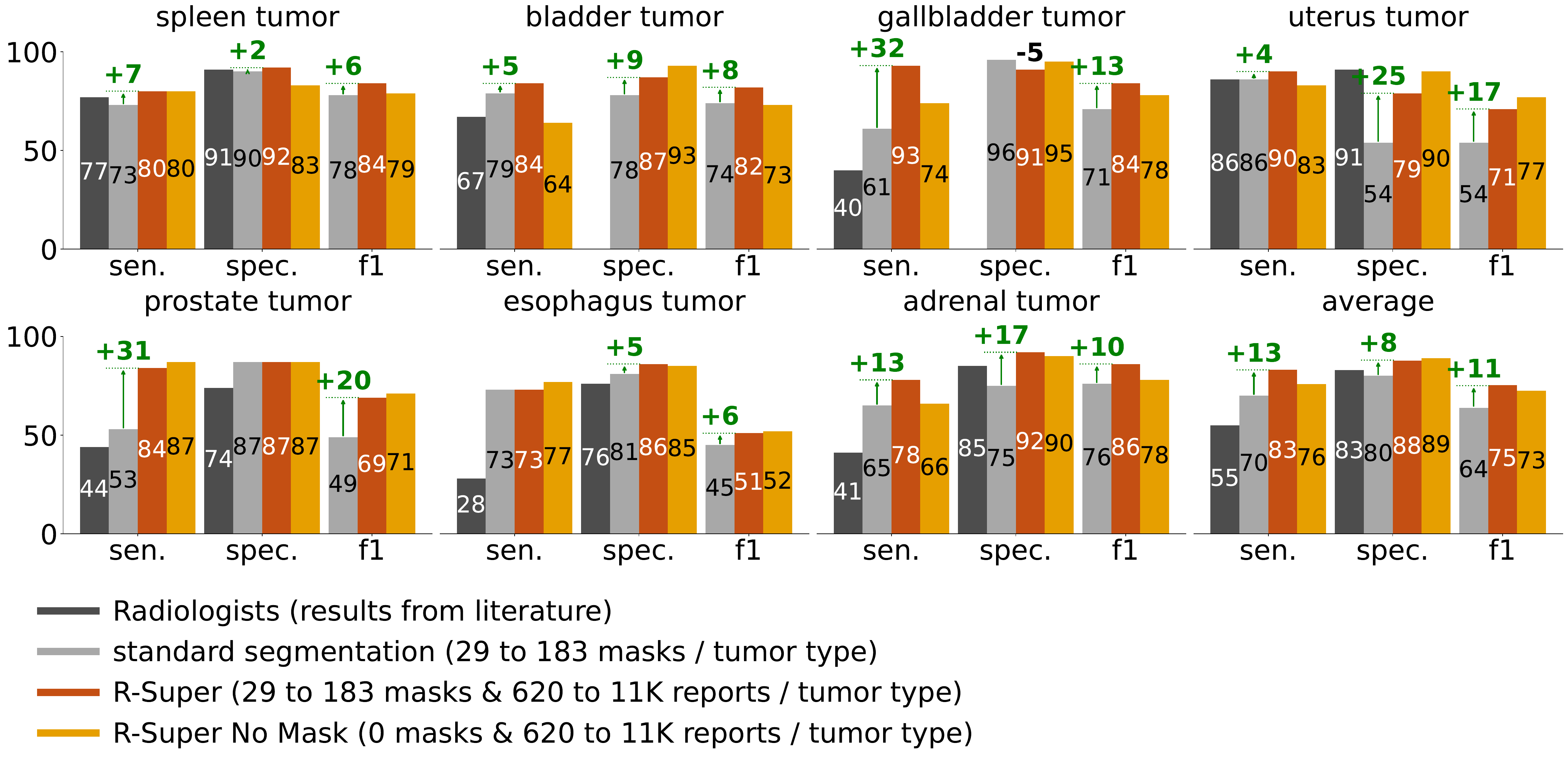}
    \caption{\textbf{In external validation, \method\ outperforms standard segmentation (trained only with CT-Mask pairs) by large margins.} \method\ surpasses radiologist tumor detection performance for six of the seven tumor types. Radiologist performance was extracted from the literature, see Appendix \ref{app:radiologist_studies} for an analysis of the selected studies and limitations of the comparison. Even when trained with CT-Report pairs (620 to 11K CT-Report pairs per tumor type) and zero CT-Mask pairs, \method\ surpassed standard segmentation (trained with 29 to 183 CT-Mask pairs per tumor type, \tableautorefname~\ref{tab:datasets_summary}). \method\ trained with both CT-Report pairs and CT-Mask pairs achieved the best results, surpassing standard segmentation by +12\% F1-Score. We test on a hospital never seen during training, the Stanford Hospital (Merlin Test Set, N=1,133). All segmentation models were trained on the UCSF dataset and tested on Merlin. Esophagus tumor F1-Score seem low due to a large unbalance in the test set: only 21 esophagus tumor cases for 170 normals. See \tableautorefname~\ref{tab:datasets_summary} for dataset details.
    %When testing for tumor segmentation, we leveraged 111 multi-tumor masks that our radiologists created for Merlin to calculate DSC and NSD. R-Super trained without masks (CT-Report only) surpassed standard segmentation in tumor detection, but not in tumor segmentation. \method\ trained with both CT-Mask and CT-Report pairs largely surpassed standard segmentation in tumor detection and segmentation. These results indicate that including a few masks in training is important to obtain accurate tumor outlines, and that including CT-Report pairs to a CT-Mask datasets can allow the segmentation model to outline tumors better.} 
    }
    \label{fig:merlin_test}
\end{figure}

%on the importance of external validation
%cite isnet
When tested on a hospital never seen during training, \method\ outperforms standard segmentation (trained without reports) by a large margin (\figureautorefname~\ref{fig:merlin_test}). External validation of medical AI on hospitals outside the training data is essential to demonstrate that the AI model can perform well across institutions, patient demographics, clinical procedures, and CT scanners \cite{bassi2024improving,geirhos2020shortcut,degrave2021ai,bassi2024touchstone}. To perform external validation, we excluded the Merlin dataset from training. We trained \method\ on all UCSF CT–Report and CT-Mask pairs, and tested only on Merlin. Merlin comes from the Stanford Hospital, which is not in the UCSF dataset---making Merlin out-of-distribution. %Here, we also evaluate \method\ with DSC and NSD, which evaluate how well AI models can outline tumors and find their borders. To calculate DSC and NSD, we compared the \method\ outputs to 111 tumor masks that our radiologists manually created for Merlin. To foster research in the domain, we will make these masks public---the first public tumor masks for the seven tumor types we address here.
%\TODO{DICE! Problem: we annotated the worst cases possible with masks, maybe pick the random selected ones for testing.}

As in internal validation (Tab. \ref{tab:multi_tumor_results}), method surpassed radiologist performance in external validation (Fig. \ref{fig:merlin_test}) for tumors in the bladder, gallbladder, uterus, prostate, esophagus, and adrenal glands. Additionally, in external validation \method\ also surpassed radiologist performance for spleen tumors. The radiologist performances were extracted from published studies, and Appendix \ref{app:radiologist_studies} describe the studies and limitations of this comparison.

\subsection{\method\ Also Improves Segmentation of Tumors with Many Masks}
\label{sec:pants}

\begin{figure}
    \centering
    \includegraphics[width=1\linewidth]{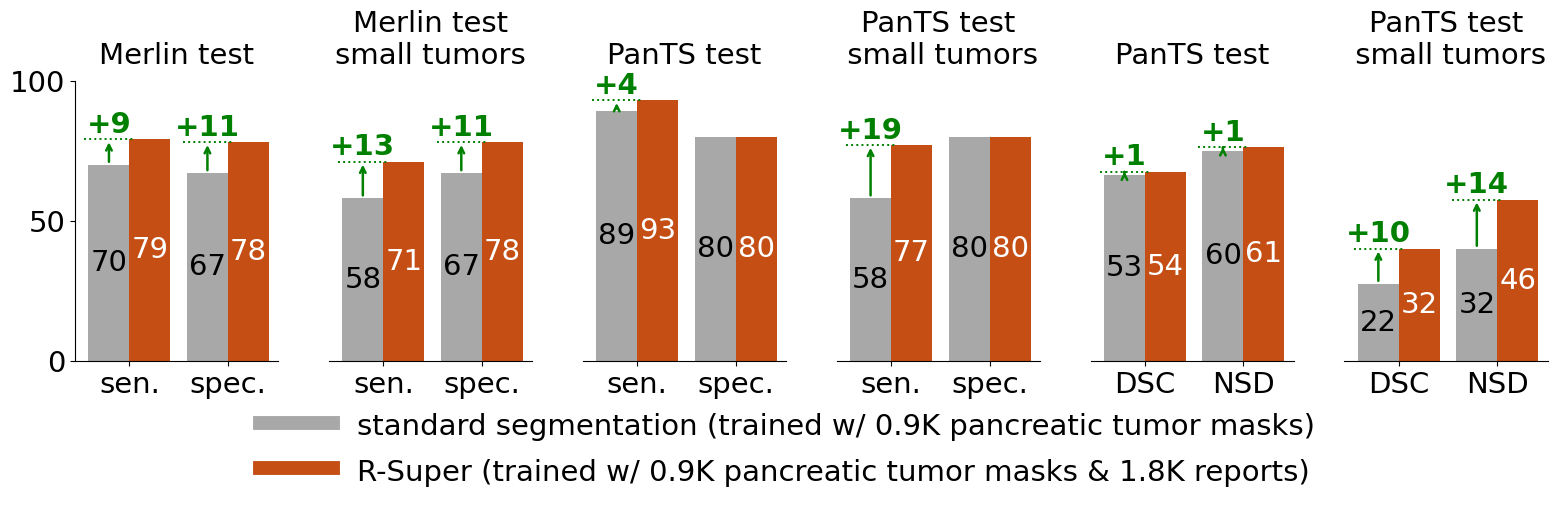}
    \caption{\textbf{\method\ scales the largest public pancreatic tumor segmentation dataset, improving AI performance---especially for small tumors ($<$ 2 cm).} PanTS \cite{li2025pants} is the largest public CT-Mask dataset for pancreatic tumor segmentation (1.1K pancreatic tumor masks). We scale it by merging PanTS and Merlin \cite{blankemeier2024merlin}, a public CT-Report dataset with 2K pancreatic tumor reports. By learning from CT-Mask and CT-Report pairs (PanTS \& Merlin), \method\ substantially outperformed a standard segmentation model, trained only on CT-Mask pairs (PanTS). We evaluated in a Merlin test set (400 CTs, 200 with pancreatic tumors), and the PanTS test set (901 CTs, 151 with pancreatic tumors). Notably, \method\ had the largest advantage in small pancreatic tumors (e.g., +19\% sensitivity for small tumors in PanTS)---critical for early detection. We evaluated both for tumor detection (sensitivity and specificity) and segmentation (DSC and NSD). DSC and NSD are only possible to calculate in PanTS, because it has ground-truth tumor masks. Also, we calculate DSC and NSD only for CT scans with tumors. For small tumors, \method\ produced a strong improvement in DSC and NSD. For larger tumors, the improvement was smaller, possibly indicating an overfit of the standard segmentation model (trained on PanTS only) to the PanTS masks.}
    \label{fig:pants_results}
\end{figure}

%show it works with public data
\method\ not only enables the segmentation of tumors when few or no masks exist (Section \ref{sec:multi_tumor_results} to \ref{sec:external_eval}), it also improves the segmentation of tumors that are the focus of the largest public CT-Mask datasets (\figureautorefname~\ref{fig:pants_results}). The PanTS Dataset \cite{li2025pants} is the largest public dataset with pancreatic tumor masks. It includes 9,000 public CT-Mask pairs, 1,077 with pancreatic tumors. Therefore, AI trained on PanTS represents the state-of-the-art of what standard segmentation training can achieve with public CT-Mask pairs. We show that, using \textit{public data only}, \method\ advances this state-of-the-art substantially. To train on public data only, we do not use our 101,654 CT-Report dataset here. Instead, we train \method\ on PanTS-train (900 pancreatic tumor CT-Mask pairs) plus Merlin-train (1,800 pancreatic tumor CT-Report pairs). \figureautorefname~\ref{fig:pants_results} shows that \method\ substantially outperformed standard segmentation (trained only on the PanTS-train CT-Mask pairs) both on the PanTS test set, and on the Merlin test set. \method\ was especially helpful for small tumors, providing up to +19\% in sensitivity at matched specificity, and +10\% DSC. Thus, \method\ can use reports to scale the largest public segmentation datasets, improving AI performance and early detection of tumors. Notably, whereas our previous experiments trained \method\ on a ratio of 100 CT–Report pairs per CT–Mask pair, this experiment used only a ratio of 2 CT-Reports pairs per CT-Mask pair—yet \method\ still yielded substantial gains. Thus, one does not need an enormous number of CT-Report pairs to benefit from \method.

\section{Discussion}

This study introduces \method, a novel AI training method that converts reports into supervision signals that directly guide the segmentation task---constraining segmented tumors to match the tumor count, diameters, volumes, attenuations, and locations in reports. We used \method\ to train on a dataset with an unprecedented number of CT-Report pairs---101,654. In five test datasets, encompassing both internal and external validation, \method\ substantially surpassed other AI training methods and state-of-the-art VLMs in detecting tumors---such as Merlin, from Stanford University, and MedGemma, from Google. By effectively learning from reports, \method\ surpassed standard segmentation training (without reports) by a very substantial margins: +13\% sensitivity, +8\% specificity and +11\% F1-Score in external validation. Additionally, \method\ can train any segmentation architecture. It does not significantly increase training time\footnote{We trained \method\ in five days with 2 NVidia H100 GPUs.}, and does not change inference time. We will release the first public AI capable of segmenting tumors in the spleen, esophagus, adrenal glands, bladder, gallbladder, uterus, and prostate, in CT scans. We plan to keep expanding the size and types of cancer in our dataset. We are contacting multiple medical institutions, in diverse countries, to collaborate with CT scans and reports. Furthermore, we are gathering other types of medical images, such as MRI---where we hypothesize \method\ could be directly applied.

This study reveals the importance of reports for tumor segmentation. By introducing a novel training method that can effectively learn tumor segmentation from reports, we demonstrated that report-based training can yield double-digit improvements in tumor detection and segmentation metrics in 3 test datasets (UCSF, Merlin, and PanTS). Our results demonstrate that reports allows tumor segmentation with few or zero tumor masks---training with many reports (101,654) and zero mask surpassed training with zero reports and few masks (723),  \tableautorefname~\ref{tab:multi_tumor_results}). Thus, \method\ allowed the segmentation of seven tumor types missing from public CT-Mask datasets. Moreover, R-Super can use CT-Report pairs to scale the largest public CT-Mask training datasets (e.g., PanTS, the largest public pancreatic tumor CT-Mask dataset) and substantially improve results. In summary, we show that reports can strongly improve tumor segmentation and detection---given an AI training method that effectively learns from reports, like \method. We hope this result can encourage more researchers to develop report-based training methods. Overall, we hope our findings will advance the tumor segmentation field, helping in a transition from small CT-Mask datasets to large CT-Mask \& CT-Report datasets.

%We believe that open science is the most efficient way to advance medical AI. Thus, 
To benefit the research community, we will make public the \method\ code, the first public AI model that can segment and detect seven understudied tumor types in CT, and the first public tumor masks for these tumor types. Until now, to segment these seven %\szymon{sometimes you write 7, sometimes 7, which is a standard here?} 
tumor types unavailable in public CT-Mask datasets, radiologists needed to spend months or years drawing tumor masks. In previous studies, eight radiologists spent five years to produce 3,125 masks for pancreatic tumors. These same radiologists would need 300 years to create masks for the 101,654 CT scans in our dataset. 
Mask creation represents an enormous cost and time barrier, which has been preventing broader research on multi-tumor segmentation. \method\ removes this barrier by allowing AI to train with CT scans and reports---readily available in hospitals and public datasets. In summary, study provides the community with data and an efficient training method to segment understudied tumor types. 
%\sd{Given the large availability of reports and well-known associated privacy issues we are also planning to provide a federated learning version of this algorithm, hence leveraging, with privacy-by-design constraints, networks of hospitals without any sharing of patients data. This will easy the application of the method}.\\
We hope this contribution to help democratize AI research and foster further advancements in the detection of understudied tumor types. In the end, we expect this research will translate into better cancer detection. 

%\szymon{TODO: Maybe it's a good idea to highlight within Discussion, that to train \method\, we only need raw data extracted from hospital database -- CT + report, which is commonly stored around the World, without further assistance or supervision from the clinicians?}

%Comparision to VLMs: reports -> precise localization cues, restinged to a few voxels in the CT, intead of just trying to align embeddings.

\section{Methods}
\subsection{Assembling a Training Dataset of 101,654 CT-Report Pairs}
\label{sec:methods_dataset}

\begin{figure}
    \centering
    \includegraphics[width=1\linewidth]{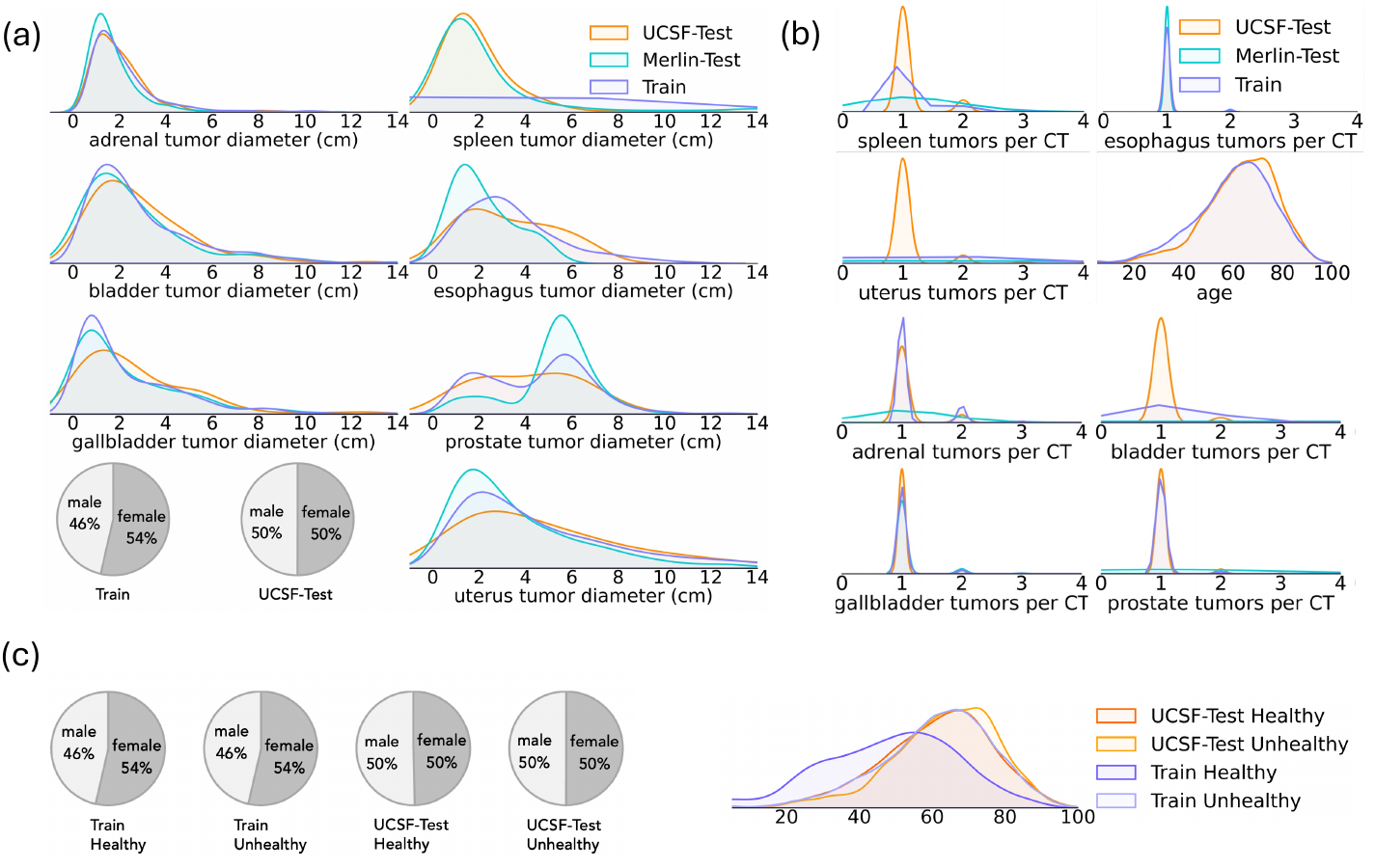}
    \caption{\textbf{Dataset summary.} \textbf{(a)} Distribution of tumor diameters and patient sex in the training dataset (UCSF \& Merlin train) and test datasets (UCSF test and Merlin test). \textbf{(b)} Distribution of tumor counts per CT scan and patient age in the training and test datasets. \textbf{(c)} Comparison of age and sex distribution for healthy and unhealthy (tumor) patients. The UCSF test set was randomly selected, matching the age and sex distribution from healthy and unhealthy patients---avoiding bias in our results.}
    \label{fig:dataset_statistics}
\end{figure}

This study is built upon three datasets: UCSF, Merlin, and PanTS.

%patient split

\textbf{(1) UCSF Dataset.} LLMs searched over 410,000 CT reports from the University of California San Francisco (UCSF) Picture Archiving and Communication System (PACS). These reports are from 1997 to 2024, and they encompass the UCSF hospital and multiple affiliated institutions in California, USA. The LLM read the reports and selected normal patients and those with tumors in the esophagus, bladder, gallbladder, spleen, uterus, prostate, and adrenal glands. For efficiency, we used a small LLM, Llama 3.1 8B AWQ. Then, a large LLM, LLama 3.1 70B AWQ, read the selected reports again, confirming the small LLM findings. The LLMs used radiologist-designed prompts, available in our public code. To certify LLM accuracy, radiologists read 447 of the reports selected by the LLM, and certified that it has 96\% accuracy in identifying patients with tumors\footnote{In the verified reports, 182 patients had tumors, 265 were normals. The LLM correctly identified all tumor reports, and correctly identified 247/265 of the normal reports.}---a level of accuracy on par with labelers in established datasets like CheXpert \cite{irvin2019chexpert} and ChestX-ray 14 \cite{wang2017chestx}. In total, the LLMs selected 85,899 reports of interest. The UCSF dataset covers the pelvis, abdomen and chest. It includes non-contrast and contrast cases---84\% are venous phase, 10\% arterial phase, and 6\% non-contrast. Of all CT scans, 68\% were for outpatients (same-day visits), 17\% for inpatients (admitted patients), and 15\% were done in the emergency department (urgent care). In total, the dataset has 33,248 patients and 85,899 CT-Report pairs. Prior to this study, all data was de-identified.

\textbf{(2) Merlin Dataset \cite{blankemeier2024merlin}.} Merlin is the largest public abdominal-focused CT-Report dataset. It was collected from the Stanford Hospital, and includes 25,494 scans from 18,317 patients, acquired from 2012 to 2018. Every CT scan is paired with its report. Exams were selected via the Stanford Medicine Research Data Repository (STARR), using Current Procedural Terminology (CPT) codes 72192, 72193, 72194, 74150, 74160, 74170, 74176, 74177, and 74178. Of the CT scans in Merlin, 97\% are portal venous, 2.4\% delayed, 0.45\% arterial, and 0.26\% non-contrast. CT scans include the abdomen, chest and pelvis. All data was de-identified by the dataset creators.

\textbf{(3) PanTS Dataset \cite{li2025pants}.} PanTS is the largest public CT-Mask dataset focused on pancreatic tumors. It includes 9,901 public CT scans from 143 medical institutions in 17 countries; including 1,077 CT-Mask pairs with pancreatic tumors. All CT scans have masks for 27 organs and anatomical structures (including the pancreas and its head, body, and tail). Contrast phases are: non-contrast 7.9\%, venous 64.9\%, arterial 26.6\%, and delayed 0.6\%.  All data was de-identified by the dataset creators.

\subsubsection{Training and Testing Splits}
\tableautorefname~\ref{tab:datasets_summary} summarizes the training and testing datasets used in each of our experiments. \figureautorefname~\ref{fig:dataset_statistics} provides tumor and demographic information on our training and testing datasets. In our main experiments (Section \ref{sec:multi_tumor_results} and Section \ref{sec:small_tumor_results}), we trained \method\ on 101,654 CT-Report pairs; 82,130 from the UCSF dataset, 25,494 from Merlin. In Section \ref{sec:multi_tumor_results}, we tested on 1,220 CT scans from unseen patients at UCSF. Test CTs were randomly selected, ensuring similar age and sex demographics between normal and tumor patients. In Section \ref{sec:small_tumor_results}, we tested on the CT scans with small tumors (213 CT scans with tumors smaller than 2 cm) and normal cases (257) inside this UCSF test dataset. In Section \ref{sec:external_eval}, for external validation, we trained on the UCSF dataset and tested on a Merlin test set. Finally, in Section \ref{sec:pants}, we trained on Merlin and PanTS, and tested on the official PanTS test split (901 CT scans, 151 with pancreatic tumor) and 400 CT scans randomly selected from Merlin (200 with pancreatic tumors, 200 normals). All training datasets (including PanTS) contain normal cases, benign tumors, and malignant tumors. The Merlin and UCSF test sets include only malignant tumors (primary or metastatic) and healthy controls. Malignancy is confirmed through explicit mentions in radiology reports or, for the UCSF test set, by pathology reports.

%training and testing split

%\TODO{Appendix: training parameters and descriptions of baselines.}

\subsection{Report-based Active Learning}
\label{sec:active_learning}

\begin{figure}
    \centering
    \includegraphics[width=0.75\linewidth]{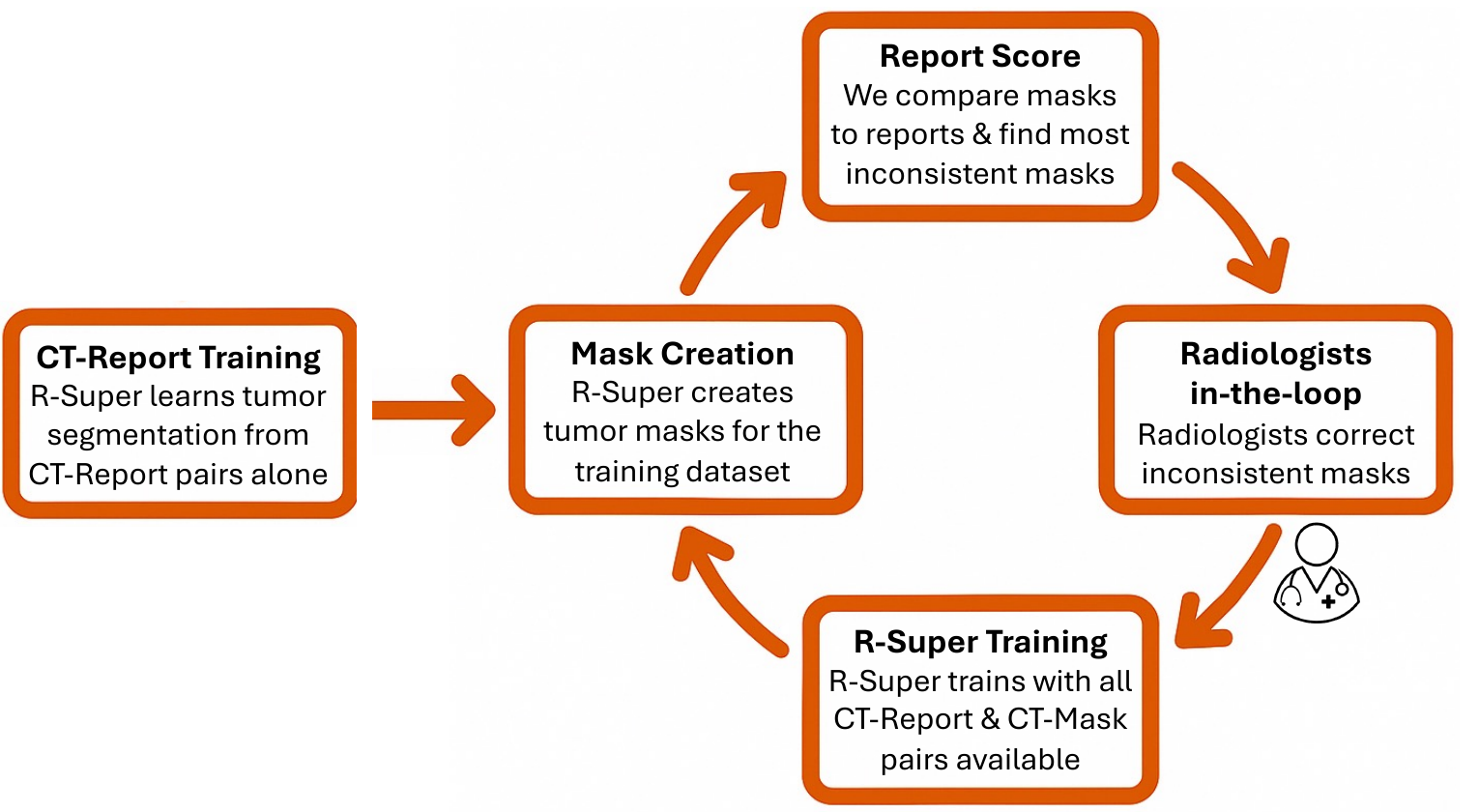}
    \caption{\textbf{Our report-based active learning enables radiologists to create tumor masks six times faster.} Instead of drawing masks from scratch, radiologists correct AI-generated masks and have access to the original radiology reports—reducing annotation time from 30 to five minutes. The process prioritizes the most inaccurate masks, identified automatically using the Ball Loss, which increases when AI predictions disagree with report details (tumor number, size, or location). Retraining on these corrected masks helps \method\ avoid previous errors. Standard active learning begins with radiologists creating masks from scratch, without AI assistance—the cold-start problem. In contrast, our strategy starts with \method, already trained from CT–Report pairs, providing AI assistance from the start. As radiologist-corrected masks are created, \method\ continuously improves and consistently outperforms standard segmentation models trained without reports, making it a more effective assistant throughout the entire active learning process.
    }
    \label{fig:active_learning}
\end{figure}

Drawing a single tumor mask in CT (which is tri-dimensional) can take up to 30 minutes for a radiologist, making large-scale mask creation costly and slow \cite{qu2023annotating,li2024abdomenatlas}. Our dataset includes 723 tumor masks, created by 31 radiologists using a new report-based active learning strategy built on \method. Radiologists reported that this strategy reduced annotation time from around 30 to five minutes per mask.

Figure~\ref{fig:active_learning} illustrates our active learning strategy. We first train \method\ on CT–Report pairs only. Then, we iteratively create tumor masks through a loop: \method\ generates AI-made tumor masks; we identify those the least consistent with reports; radiologists correct them; and we re-train \method\ with the radiologist-corrected tumor masks plus the remaining CT-Report pairs. We stopped at 723 radiologist-corrected tumor masks. Consistency between AI-made tumor masks and reports is quantified using the Ball Loss (Section~\ref{sec:ball_loss}), which increases when the AI-made tumor mask disagrees with tumor descriptions in reports—in tumor number, size, and location. We prioritize the most inaccurate AI-made tumor masks for correction. Retraining \method\ on corrected masks teaches it to avoid its past mistakes.

Unlike traditional active learning—where radiologists begin by creating masks without AI assistance—our strategy starts with a strong segmentation model to assist radiologists, \method\ trained from CT–Report pairs alone. It provides effective AI assistance from the start, accelerating early mask creation. As more radiologist-corrected masks are added to the training dataset, \method\ continuously improves. It continuously maintains superior accuracy to standard segmentation models trained without reports—throughout the whole active learning process, whether few or many masks are available (Sections~\ref{sec:multi_tumor_results}, \ref{sec:pants}). Overall, this report-guided active learning speeds up tumor mask creation by sixfold and delivers a more accurate, continuously improving AI assistance for radiologists.

\subsection{\method}

\figureautorefname~\ref{fig:r_super} is an overview of the \method\ training method, designed to enforce consistency between AI-segmented tumors and tumor descriptions in reports. Section \ref{sec:llm} explains how \method\ uses an LLM to extract tumor characteristics from reports---tumor count, locations\footnote{Locations refer to the organ or organ sub-segment where the tumor is. In our experiments, we used sub-segments for pancreatic tumors (pancreatic head, body, or tail), and organs for other tumors. When available, we also extract the tumor slice---the horizontal plane where the tumor is in the CT.}, diameters, attenuation, and estimated volumes. Section \ref{sec:volume_loss}, Section \ref{sec:ball_loss}, and Section \ref{sec:att_loss} explain the three novel loss functions that use the LLM-extracted tumor characteristics as ground-truth: the Volume Loss, Ball Loss and Attenuation Loss, respectively. The Volume Loss, used as deep supervision, is less strict, enforcing only volume and location consistency between the segmented tumors and reports. The Ball Loss directly supervises the final AI segmentation output, and it enforces consistency in tumor volumes, locations, count, and diameters. The Attenuation Loss is applied both as deep supervision and at the output. It enforces consistency in attenuation---if a report states that a tumor is hypoattenuating, the segmented tumor should be darker than the surrounding organ; if hyperattenuating, it should be brighter. 

\subsubsection{LLM for Extracting Report Information}
\label{sec:llm}

We use an LLM to extract tumor characteristics from reports. It directly extracts tumor counts, locations (organ/organ sub-segment/tumor slice), attenuation and diameters. Tumor volumes are estimated from diameters (see Eq. \ref{eq:vol_loss}). Since LLMs can interpret semantics and context, they can adapt to the diverse writing styles and word choice of reports---written in diverse medical institutions by diverse radiologists. 
To facilitate the application of \method\ to any hospital and avoid any risk of overfitting the LLM to the styles of the reports in our training dataset, we use a zero-shot LLM, Llama 3.1 70B AWQ \cite{touvron2023llama}. Notably, zero-shot LLMs extract tumor characteristics from reports accurately, according to manual evaluation by radiologists (Section \ref{sec:methods_dataset}). To reduce computational cost, we run the LLM only once per report and store its answer. Our LLM prompt (available in our code) was designed by radiologists in an iterative procedure\footnote{Radiologists and computer scientists prepared a prompt, tried it, checked for LLM errors, and improved the prompt to avoid these errors.}. This prompt provides the LLM with medical knowledge and detailed guidelines for understanding reports. The prompt also asks the LLM to thoroughly justify its answers according to the report, and to fill templates with the tumor characteristics (tumor diameters, organ, organ sub-segment, slice, attenuation). The LLM-filled templates are automatically converted into a table for later use as ground-truth for the Volume Loss, Ball Loss, and Attenuation Loss. Sometimes, reports miss some tumor characteristics (e.g., diameters). Our loss functions also work for these reports---they leverage all information \textit{available} in each report (Sections \ref{sec:volume_loss} and \ref{sec:ball_loss}).

\subsubsection{Volume Loss}
\label{sec:volume_loss}

We apply the Volume Loss to a deep layer of the segmentation model, as deep supervision\footnote{Before applying the loss, we use a $1 \times 1 \times 1$ convolution with sigmoid activation to reduce the number of channels in the deep layer output, making them match the number of segmentation classes. We also use nearest neighbor interpolation to make the deep layer output match the input size and voxel spacing.}. The Volume Loss is designed to be \textit{not} strict---it enforces only two constraints: tumors must be segmented inside the \textbf{locations} (organs or organ sub-segments, for simplicity we say ``organ'' in the explanations below) where the report mentions tumors, and the \textbf{combined volume} of all segmented tumors must match the combined volume of all reported tumors in each location. The non-strict loss allows exploration in deeper layers, while the strict Ball Loss (see Section \ref{sec:ball_loss}) enforces accurate final predictions. When reports inform the tumor slices (the vertical height of the tumor in the CT), the Volume Loss also enforces the segmented tumors to be at the informed slices.

Usually, reports do not directly provide tumor volumes, but they provide tumor diameters. We extract diameters with the LLM and use them to estimate volumes. Reports can provide 1, 2, or 3 diameters for a tumor\footnote{One diameter measurements are common for small, rounder tumors, and they are used in the RECIST (Response Evaluation Criteria in Solid Tumors) guideline \cite{eisenhauer2009new}. Two diameters are used in the World Health Organization (WHO) tumor measurement standard \cite{miller1981reporting}, where the first diameter is the largest tumor diameter in any CT axial slice, and the second diameter is measured perpendicularly to the first, in the same slice. Some reports have a third diameter, perpendicular to the other two.}. With a single diameter ($d_1$), we estimate tumor volume as a ball: $d_1^3 \pi/ 6$. With 3 diameters ($d_1$, $d_2$, $d_3$), we use an ellipsoid estimation: $d_1d_2d_3\pi/ 6$. With 2 diameters, $d_1$ and $d_2$, we estimate $d_3$ as $(d_1+d_2)/2$, and use the ellipsoid volume estimation. After estimating the volume for each tumor the report describes in an organ $o$, we sum them, giving $V_{r,o}$---the \textit{total reported tumor volume in $o$}.

The Volume Loss optimizes the total \textit{segmented} tumor volume, $V_{s,o}$, to match $V_{r,o}$, for each organ $o$. To calculate $V_{s,o}$, we divide the CT into organs (and organ sub-segments) using pre-saved, AI-made organ masks. These organ masks do not need to be manually created. We created them with an nnU-Net \cite{isensee2021nnu} trained on public data\footnote{Organ segmentation is usually more accurate than tumor segmentation---DSC scores above 80\% are common in organ segmentation \cite{bassi2024touchstone}, but state-of-the-art tumor segmentation models rarely reach 70\% DSC \cite{chen2023cancerunit,plotka2025mamba}. Public CT datasets have few masks for few types of tumors, but these datasets have many masks for many organs \cite{li2024abdomenatlas,wasserthal2023totalsegmentator}. Moreover, there are many accurate and public organ segmentation models, such as TotalSegmentator \cite{wasserthal2023totalsegmentator} and Touchstone \cite{bassi2024touchstone}. Our organ segmentation model was trained on the public dataset AbdomenAtlas \cite{bassi2025radgpt}. It segments 39 organs and structures, including all seven organs where our dataset has tumors, and pancreas sub-segments.}, and we will also publicly release this nnU-Net. To compensate for errors in organ masks and account for tumors that grow beyond organ boundaries, we expand the organ masks with binary dilation (by about 2 cm). When tumor slices are informed in the report, we just edit the organ mask, $\bm{O}=[o_{h,w,l}]$, making it zero in regions away from the informed tumor slices\footnote{Consider tumor slices $z_{i}$, for tumors i with maximum diameter $d_{i}$; z is the vertical axis of the CT. We make zero the organ mask in all z coordinates where the distance to any tumor slice $z_{i}$ is larger than the corresponding tumor diameter $d_{i}$. I.e., $o_{h,w,l}=0$ if $|h-z_{i}|>d_{i}$ for all tumors i.}. The Volume Loss will automatically discourage tumor segmentations in these zeroed regions. Then, for each organ with tumors in the report, $o$, we multiply (element-wise) the organ mask, $\bm{O}=[o_{h,w,l}]$, with the tumors segmented by the AI for the organ $o$ (after softmax/sigmoid activation function), $\bm{T}^{o}=[t_{h,w,l}^{o}]$. The multiplication selects only the tumors segmented inside the organ $o$. To estimate the total tumor volume in $o$, $V_{s,o}$, we sum the multiplication result in the spatial dimensions, and multiply it by the volume of one voxel, $v$:

\begin{equation}
\label{eq:vol_calculation}
    V_{s,o} = v \sum_{h,w,l}^{H,W,L} t_{h,w,l}^{o} o_{h,w,l}
\end{equation}

The Volume Loss minimizes the difference between $V_{s,o}$ and $V_{r,o}$ (ground-truth). To this end, we experimented with many loss functions, like the L1 and L2 losses, but we achieved better convergence with the function in Eq. \ref{eq:loss_formula}. The  reasons for this better convergence are: (1) a strong but finite gradient when $V_{s,o}=0$ and $V_{r,o}\neq0$, strongly penalizing the AI when it misses tumors, but keeping numerical stability; (2) a soft gradient when $V_{s,o}>V_{r,o}$, since a strong gradient when $V_{s,o}>V_{r,o}$ can increase the number of tumors missed by the AI (by pushing it towards a $V_{s,o}=0$ solution).

\begin{equation}
\label{eq:loss_formula}
L_{\text{forg},o}'(V_{s,o},V_{r,o}) = \frac{|V_{s,o} - V_{r,o}|}{V_{s,o} + V_{r,o} + E}
\end{equation}

\begin{figure}[t]
    \centering
    \includegraphics[width=1\linewidth]{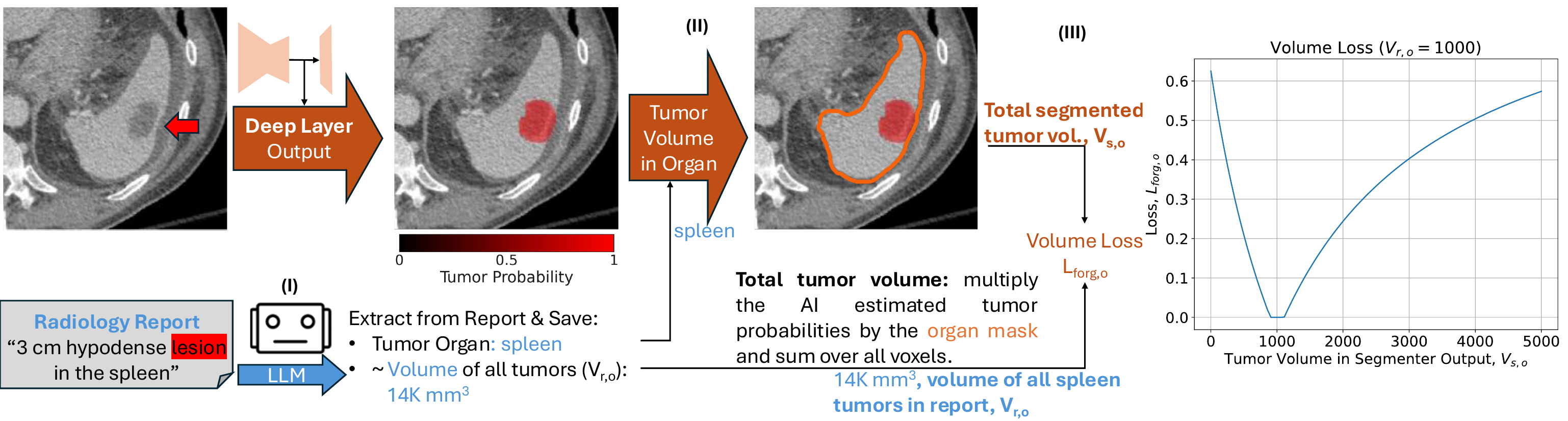}
    \caption{\textbf{The Volume Loss enforces the volume of segmented tumors to match the tumor volume estimated from the report.} The loss is applied to deep layers of the segmentation model, as deep supervision. \textbf{(I)} An LLM extracts tumor diameters and locations (organ/organ sub-segment/slice) from reports. From diameters, we estimate the total tumor volume from the radiology reports, $V_{r,o}$, within each organ/organ sub-segment $o$. \textbf{(II)} We sum the AI tumor segmentation output (softmax/sigmoid) for all voxels inside the organ/sub-segment $o$, estimating the total segmented tumor volume in the organ/sub-segment, $V_{s,o}$. Pre-saved, AI-made organ/sub-segment masks identify the voxels inside the organ/sub-segment. \textbf{(III)} We use a custom regression loss (Eq. \ref{eq:vol_loss}, right panel of the figure) to enforce the segmented tumor volume, $V_{s,o}$, to match the tumor volume in the radiology report, $V_{r,o}$. This loss includes a tolerance margin to account for human and estimation errors of $V_{r,o}$. The figure plots the loss for $V_{r,o}=1000$ mm$^3$ and varying $V_{s,o}$. In case the report informs the tumor slices, the Volume Loss also ensures that the segmented tumors are near the informed slices.}
    \label{fig:volume_loss}
\end{figure}

The constant $E$ (set to 500 mm$^{3}$) provides numerical stability for small $V_{r,o}$. Importantly, the tumor volumes estimated from reports ($V_{r,o}$) are not perfect. They are subject to human errors, inter-observer variance, and approximation errors in our volume estimation from diameters. Thus, we added a \textbf{tolerance margin} ($0<\tau<1$) in the Volume Loss: if the difference between $V_{s,o}$ and $V_{r,o}$ is small (i.e., $|V_{s,o}-V_{r,o}|\leq\tau V_{r,o}$) the Volume Loss does not penalize the AI---the loss and its gradient become zero. Eq.~\ref{eq:vol_loss} displays the loss with tolerance, and \figureautorefname~\ref{fig:volume_loss} plots it. We set $\tau=10\%$.

%\szymon{Before below equation, there should be a some introduction sentences}

\begin{equation}
\label{eq:vol_loss}
L_{\text{forg},o}(V_{s,o},V_{r,o}) = \max\{L_{\text{forg},o}'(V_{s,o},V_{r,o})-L_{\text{forg},o}'((1-\tau) V_{r,o},V_{r,o}),0\}
\end{equation}

For each organ $o$ with tumors in the report, the Volume Loss also penalizes any tumor segmented outside the organ, using cross-entropy and the organ segmentation mask (Eq. \ref{eq:background_tumor}). For organs with no tumor in the report, we use cross-entropy to penalize all tumor segmentation output voxels, pushing them towards 0. Equation \ref{eq:vol_final} displays the final Volume Loss: $L_{\text{forg},o}(V_{s,o},V_{r,o})$ makes $V_{s,o}$ match $V_{r,o}$ inside the organ $o$ with tumors, and the term $L_{\text{bkg},o}(\mathbf{T}^{o})$ minimizes tumor segmentation outside this organ.

%\szymon{Here as well, sentences before these equations.}

\begin{gather}
\label{eq:background_tumor}
L_{\text{bkg},o}(\mathbf{T}^{o}) = -\frac{1}{H \cdot W \cdot L} \sum_{h=1}^{H} \sum_{w=1}^{W} \sum_{l=1}^{L} \ln(1 - t_{h,w,l}^{o}(1-o_{h,w,l})) \\
\label{eq:vol_final}
L_{\text{vol},o} = L_{\text{forg},o}(V_{s,o},V_{r,o}) + L_{\text{bkg},o}(\mathbf{T}^{o})
\end{gather}

Following its non-strict design, the Volume Loss allows flexibility in how tumors are segmented inside organs. Also, it does not push estimated tumor probabilities towards 1 or 0, allowing uncertainty and exploration in deep layers. Mainly, the Volume Loss enforces that the AI must \textbf{not}: (I) segment tumors in organs the report mentions no tumor (false-positive), (II) miss tumors in organs the report mentions tumors (false-negative), (III) segment tumors much larger/smaller than tumors in the report, (IV) segment tumors away from informed tumor slices (when reports inform them).

When a report mentions that an organ has tumors, but it does not inform tumor diameter or number of tumors, we cannot estimate the total reported tumor volume in the organ, $V_{r,o}$. In this case, we resort to a prior-based, high-tolerance version of the tumor loss: we consider that the tumor must be larger than 5 mm in diameter ($V_{r,o}>65$) and smaller than 120 mm ($V_{r,o}<904,779$). These numbers are based on the analysis of tumor sizes in our dataset. To implement this requirement, when the real $V_{r,o}$ is unknown (i.e., report does not inform tumor number or size), we substitute $V_{r,o}$ by $\widehat{V}_{r,o}$ and calculate the loss as defined below:

\begin{gather}
\widehat{V}_{r,o} =
\begin{cases}
65 & \text{if } V_{s,o} < 65,\\[4pt]
V_{s,o} & \text{if } 65 \le V_{s,o} \le 904{,}779,\\[4pt]
904{,}779 & \text{if } V_{s,o} > 904{,}779,
\end{cases}
\\[6pt]
L_{\text{vol},o}
= L_{\text{forg},o}\!\left(V_{s,o},\,\widehat{V}_{r,o}\right)
+ L_{\text{bkg},o}\!\left(\mathbf{T}^{o}\right).
\end{gather}

%\TODO{Explain the use all data thing in the Ball Loss}
\subsubsection{Ball Loss}
\label{sec:ball_loss}

\begin{figure}[t]
    \centering
    \includegraphics[width=1\linewidth]{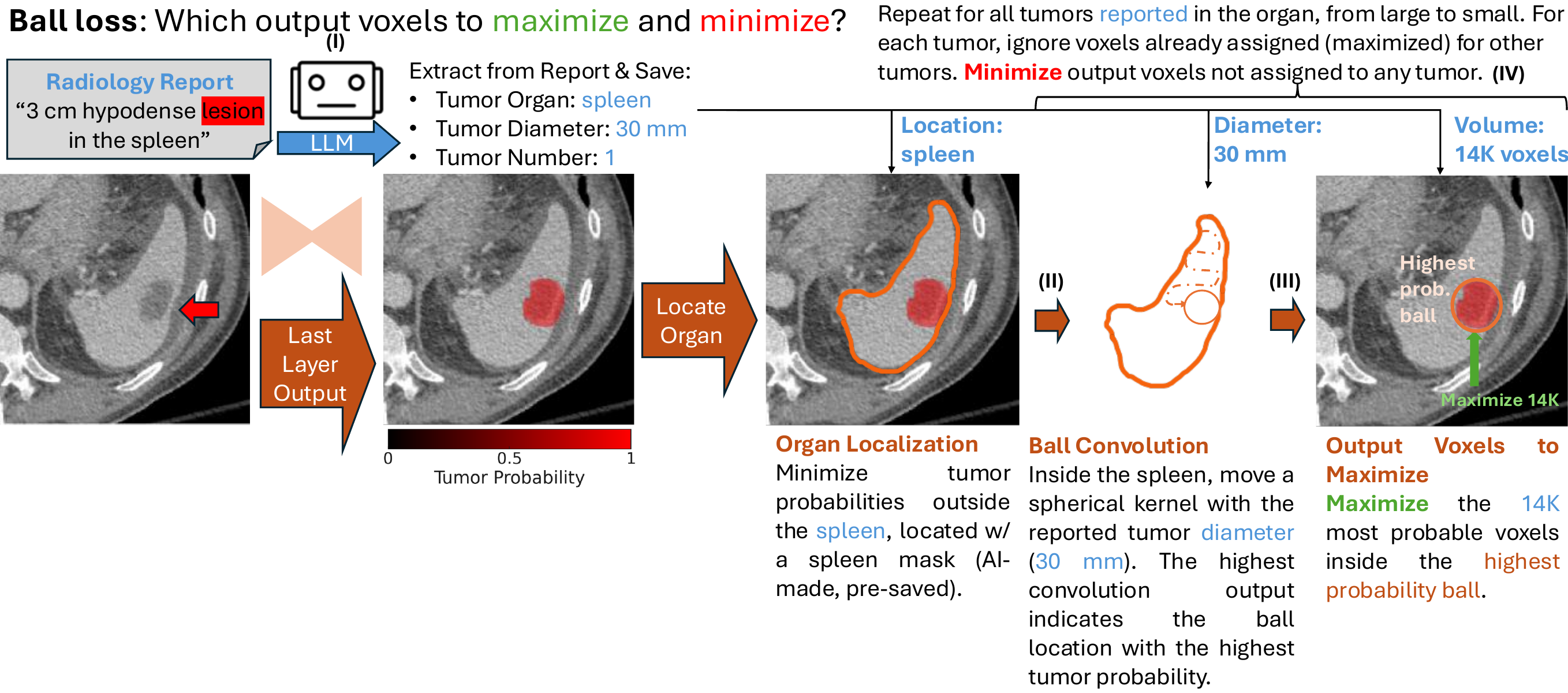}
    \caption{\textbf{The Ball Loss converts reports into voxel-level supervision.} \textbf{(I)} a zero-shot large language model (LLM) extracts and saves tumor count, locations (organs or sub-regions), slices, attenuation, and diameters from reports. 
    \textbf{(II)} During segmentation training, the tumors in the reports are located to the CT by Ball Convolutions---standard convolutions with fixed, spherical binary kernels matching reported tumor diameter (plus a small margin). We apply the convolution to the segmentation model's outputs (tumor probabilities), ignoring locations outside the organ/organ sub-segment containing the tumor (with the help of pre-saved, AI-made organ masks). When the tumor slice is informed by the report, we also ignore locations far from the slice. The convolution output is maximum when the ball is at the most probable location for the tumor---the highest probability ball.
    \textbf{(III)} By setting the top-N most probable voxels inside the highest-probability ball as 1, and the remaining voxels as 0, we create a segmentation mask for the tumor. N is the number of voxels the tumor is expected to occupy---according to its volume, estimated from the report.
    If the report shows multiple tumors, we use a sequence of ball convolutions to locate them one-by-one. After each tumor is located and added to the segmentation mask, we remove it from the segmentation output, to avoid reuse.
    \textbf{(IV)} We use the mask to optimize the segmentation model, using the dice loss and a custom cross-entropy loss, which has higher weights near the tumor center.
    The masks created by the Ball Loss have tumors with correct tumor size, count and locations (organs/sub-segments), and they get better as the segmentation model trains and improves. The Ball Loss includes tolerances for uncertain tumor borders.
    }%Editabe: https://livejohnshopkins-my.sharepoint.com/:p:/g/personal/psalvad2_jh_edu/EUqehqYjV5dAmNurdRAeLKkBUQynvsy1aMhUfB1nmaMYqw?e=VepxkO
    \label{fig:ball_loss}
\end{figure}

%\TODO{Clarify Ball Loss does not make tumors spherical}

Unlike the Volume Loss, the Ball Loss is applied to the final segmentation output of the segmentation model (last layer), and it enforces strict constraints. First, it enforces that segmented tumors must be in the \textbf{locations} (organs/organ sub-segments) where the report mentions tumors. Then, \textit{for each location}, it enforces that: the \textbf{number} of segmented tumors must match the number of tumors in the report; and each one of these segmented tumors must match the \textbf{diameter}, \textbf{volume}  and \textbf{slice} (when informed) of one tumor described in the report. Overall, the Ball Loss uses multiple information from reports to guide tumor segmentation.

First, like the Volume Loss, the Ball Loss uses the cross-entropy loss to penalize tumor segmentations for organs with no tumor in the reports. For organs with tumors in the report, \figureautorefname~\ref{fig:ball_loss} displays the Ball Loss procedure. First, we multiply (element-wise) the tumor segmentation output ($\mathbf{T}^{o}$) with the corresponding pre-saved organ segmentation mask, $\mathbf{T}^{o}\otimes\mathbf{O}$. This multiplication selects only the tumors inside the organ. Second, we apply sequential \textit{ball convolutions} to locate each individual tumor the report mentions, starting by the largest. A ball convolution is a standard convolution using a non-learnable binary kernel shaped like a ball, with the same diameter as the tumor diameter in the report (for tumors with multiple diameters, we use the largest). The convolution moves the ball inside the tumor segmentation output. In each ball position, the convolution output is the sum of all tumor probabilities (softmax/sigmoid) within the ball. The position where this sum is highest---\textit{highest probability ball}---indicates the most likely location for the tumor with the same diameter as the ball. The preliminary multiplication between the tumor segmentation output and the organ segmentation mask, $\mathbf{T}^{o}\otimes\mathbf{O}$, avoids the highest probability balls to fall outside the organ. If the report informs the tumor slice, we modify the organ mask,  $\mathbf{O}$, by making it zero on CT slices away from the informed tumor slice (more than 1 tumor diameter away). This forces the ball convolution to find a highest probability ball that intersects the informed tumor slice. Additionally, to improve tumor localization, we weigh the ball convolution kernel slightly higher around the ball center, so the convolution responds more strongly if the tumor’s center (where probabilities tend to be highest) aligns with the ball kernel’s center\footnote{Ball kernels are 1 at the kernel center, and decay towards the ball border, following a 3D Gaussian with standard deviation of $0.75 \times$ the ball diameter, $d_i$. Outside of the ball, the kernel is zero. Ball convolutions use stride 1, zero padding and odd kernel sizes to ensure input-output alignment.}.

Our first ball convolution uses the diameter of the largest tumor reported in the organ, $d_0$. The convolution output is a 3D volume, whose maximum is the most likely tumor center, $\mathbf{c_0}$, for a tumor of diameter $d_0$. We place the highest probability ball (diameter $d_0$, center $\mathbf{c_0}$) in the tumor segmentation output, and select the $N_0$ voxels with the highest tumor probability inside the ball. $N_0$ represents how many voxels the tumor should have---we derive $N_0$ from the tumor volume estimated from the report (see Section \ref{sec:volume_loss}). Then, we create an empty segmentation mask (all zero) and set these $N_0$ voxels to 1. Before using another ball convolution to locate the next tumor, we zero out these $N_0$ voxels inside the tumor segmentation output. This ensures that the next ball convolution will not locate the tumor we just located\footnote{It is important to iterate from largest to smallest tumor. Consider a segmentation output with a big tumor and a small tumor. A ball convolution with small diameter can have high outputs over the small tumor or anywhere inside the large tumor. Thus, before localizing the small tumor, we first localize the large tumor and remove it from the segmentation output.}. We repeat this process---ball convolution, add tumor to segmentation mask, remove tumor from segmentation output---until all tumors mentioned in the report are added to the segmentation mask. The final segmentation mask matches the report in tumor count, locations (organ), volumes, and diameters. We use it as ground truth for a Dice loss and a custom weighted cross-entropy loss, which optimize the AI tumor segmentation output to match the mask. Our cross-entropy gives higher weights to voxels where the AI has higher tumor confidence, and it does not penalize a margin around the tumor borders---compensating for errors in diameters in the report.

In case the report mentions a tumor but does not provide its size, we use a relaxed version of the Ball Loss. We assume that the tumor has at least 5 mm in diameter, because less than 0.1\% of tumors in our reports have less than 5 mm. Then, we apply the ball convolutions as before, considering 5 mm diameter. This will generate a small tumor inside the segmentation mask, possibly near the center of a real, larger tumor (usually centers have higher tumor probability). We use this segmentation mask to train the segmentation model with the cross entropy and dice loss. However, we do not apply the cross entropy and dice loss to the mask’s zero voxels inside the organ with tumor. The reason is that we do not know the real tumor size. Thus, it is not possible to estimate how many voxels are tumor voxels. We just enforce that, at least, a 5 mm tumor exists, but we do not penalize the segmentation model if it finds a larger tumor. In case we do not know how many tumors an organ has, we use the ball loss to localize and create a mask for each tumor described in the report, but we again do not penalize the mask’s zero voxels inside the organ with tumor.

Not all reports are equal, but the Ball Loss (like the Volume Loss) adapts to different types of reports: when reports are more precise, the Ball Loss is more precise, leveraging all available information. The most precise reports include size and slice for all tumors (about 15\% of our reports). This limits the ball convolution to search for the tumor in a very small region---a few slices inside one organ---making it very precise. 

Despite its name, the Ball Loss does not assume that tumors are spherical, nor does it teach the segmentation model to segment spherical tumors only. Instead, it assumes that tumors fit inside a ball whose diameter matches the largest tumor diameter in the report. Tumors can assume any shape that fits inside this ball. This assumption can be relaxed by increasing the diameter used in the Ball Loss (e.g., we increase the diameters in the reports by 30\% when applying the Ball Loss).

One may wonder whether the Ball Loss can guide the segmentation model to segment the wrong thing. Indeed, in a single image it can: if the segmentation model segments a wrong tumor (false positive) inside the right organ, the Ball Loss may enforce this wrong segmentation. However, when a similar false positive appears in a healthy patient, the loss penalizes it. Thus, while some wrong tumor segmentations may be reinforced in individual cases, they are canceled out across patients. To consistently minimize Ball Loss across the whole training dataset, the segmentation model must learn to find the correct tumor organs, tumor counts and tumor sizes. In other words, the only way to minimize the Ball Loss is to truly segment the tumors. Our experiments show that, by minimizing the Ball Loss and Volume Loss, R-Super becomes substantially better at segmenting tumors---proving that the net effect of our losses is reinforcing correct segmentations, not false positives.

\subsubsection{Attenuation Loss}
\label{sec:att_loss}

Reports commonly inform tumor attenuation. A hypoattenuating tumor is darker than the surrounding organ, a hyperattenuating tumor is brighter, and an isoattenuating tumor has a similar brightness to the organ. The Attenuation Loss leverages this information to improve tumor segmentation. Brightness in CT scans is expressed as HU values---the value of the voxels in the CT. Even after we normalize the CT, relative attenuation remains: if a tumor is hypoattenuating, its average voxel value will be lower than the average voxel value of the organ. We apply the Attenuation Loss both as deep supervision and at the segmentation model's final output.

To calculate the Attenuation Loss, we define the tumor voxels as the voxels where the segmentation model predicts more than 50\% tumor probability. We define the organ voxels using pre-saved, AI-made organ mask, but we do not consider tumor voxels as organ voxels. We calculate the mean and standard deviation of the tumor voxels and of the organ voxels. These means and standard deviations are sent to an MLP\footnote{128 neurons in the hidden layer.}, which is an attenuation classifier. It classifies whether tumors are all hyperattenuating, all hypoattenuating, or of mixed attenuation/isoattenuation. The label for the attenuation classifier is extracted from the report, by the LLM. We train the attenuation classifier with a standard cross-entropy classification loss. The gradient of this loss is used to train the attenuation classifier, but it also back-propagates to the segmentation model, and improves it---the segmentation model should delineate tumors better, to allow the attenuation classifier to predict the tumor attenuation better.

\backmatter

\bmhead{Acknowledgments}
This work was supported by the National Institutes of Health (NIH) under Award Number R01EB037669, the Lustgarten Foundation for Pancreatic Cancer Research, and the Center for Biomolecular Nanotechnologies, Istituto Italiano di Tecnologia (73010, Arnesano, LE, Italy). We would like to thank the Johns Hopkins Research IT team in \href{https://researchit.jhu.edu/}{IT@JH} for their support and infrastructure resources where some of these analyses were conducted, especially \href{https://researchit.jhu.edu/research-hpc/}{DISCOVERY HPC}; thank the HPC infrastructure and the Support Team at Fondazione Istituto Italiano di Tecnologia. We thank Jaimie Patterson for writing a \href{https://www.cs.jhu.edu/news/for-ai-tumor-detection-a-picture-isnt-always-worth-a-thousand-words/}{news article} about this project.

\begin{appendices}

\section{Training Details}\label{secA1}

We train using CT patches. For CT-Report pairs, each training patch is designed to fully cover one target organ. This target organ is randomly chosen, with a high probability of choosing organs with tumors in the report (e.g., 90\%). The training patch must fully cover the target organ.
Otherwise, a tumor mentioned in the report could fall outside the patch, and the report-based losses would wrongly push the AI to find a tumor that is not visible to the AI.

Training parameters followed the defaults set by MedFormer \cite{gao2022data}, the segmentation architecture we used inside R-Super. The only new parameters we include are loss weights. We set a loss weight of 1 to the segmentation losses (cross entropy and dice, used for CT-Mask pairs), 0.1 for the Volume Loss and Ball Loss, and 0.01 for the Attenuation Loss. There is no need to carefully tune these weights, we used the same weights in all our experiments. We train with AdamW, gradient norm clipping (1), 50 epochs of 1000 batches each, batch size of 4, patch size of 128 x 128 x 128, isotropic voxel spacing of 1 mm, weight decay of 5.00E-2, learning rate of 1.00E-4 (5 epochs of warmup, followed by polynomial decay). CT intensity was clipped between -991 and 500 HU, then normalized. Data augmentation includes rotation, brightness, gamma, contrast, gaussian blur and gaussian noise \cite{gao2022data}. We super-sampled the CT-Mask pairs, making them 50\% of the samples that the segmentation model saw in each epoch.

Segmentation models were initialized pre-trained for organ segmentation on AbdomenAtlas 2.0 \cite{li2024well, bassi2025radgpt}. Pre-training followed the same configuration as training (described above), but without using the R-Super losses or reports.

\section{Comparison to Radiologists}
\label{app:radiologist_studies}

The radiologist tumor detection performance is not very high for many tumor types in this study, because these tumor types are very difficult to detect in CT scans. Due to this difficulty, CT is not the primary diagnostic tool for tumors in the bladder, esophagus, prostate, uterus and gallbladder. However, more than 300 million CT scans are performed early in the world, for diverse reasons. This large number of CT scans create a large opportunity for opportunistic early detection of tumors (when tumors are found in CT scans performed for other reason, not to search for tumors). AI can help this opportunistic detection, because it can often see tumor signs that are not visible to humans. For example, PANDA detects pancreatic tumors on non-contrast CT that radiologists typically cannot \cite{cao2023large}. Here, we compared the performance of our AI to that of radiologists reported in the literature.

We searched for studies where radiologists analyzed a dataset of patients with malignant tumors and normal patients. We extracted from these studies the sensitivity and specificity reported for the radiologists. For the bladder and gallbladder, we could only find studies without healthy patients. Therefore, we report only the radiologist sensitivity in these cases. A limitation of our comparison between radiologists and AI is that our AI and the radiologists were evaluated in different test sets, with different patient populations, different CT scanners, possibly different proportions of the types of malignant tumors, different contrast protocols and different hospitals. Some clear differences are: for esophagus tumor, the radiologist study worked on non-contrast CT, while our AI worked on contrast-enhanced CT (easier); and for adrenal tumors, our AI evaluated both primary and metastatic tumors, while the radiologist study worked only on metastatic tumors. Besides the esophagus study, other studies used contrast-enhanced CT, as we did in our test datasets.
Here, we provide a brief summary of each study.

\begin{itemize}
    \item \textbf{Bladder tumors} \cite{malik2023systematic}: CT scans were selected for patients later diagnosed with bladder cancer (99 patients; 226 CTs). These scans were acquired up to five years before the pathologic diagnosis (pre-diagnostic). Radiologists achieved 67\% tumor detection sensitivity. The study lacked normal patients, so specificity was not estimable. Since these CT scans are pre-diagnostic, some may have a very small tumor, or truly no tumor, reducing the reported radiologist sensitivity.
    \item \textbf{Esophagus tumors} \cite{sui2021detection}: Non-contrast CT scans come form 52 esophagus cancer patients and 48 normal patients. Radiologists achieved 25–31\% sensitivity at 74–78\% specificity. Unlike this study, our test dataset used contrast-enhanced CT (easier). 
    \item \textbf{Gallbladder tumors} \cite{frezza1997gallbladder}: This study, published in 1997, is a retrospective analysis. It covers gallbladder carcinoma patients at the Howard University, for the previous 28 years. Radiologist performance for gallbladder tumor detection in CT was reported as 40\% sensitivity. No normal patient was included, and specificity is not reported.
    \item \textbf{Prostate tumors} \cite{korevaar2021incidental}: The study included 139 clinically significant prostate cancer CTs and 432 healthy CTs. Radiologists achieved 44\% sensitivity and 74\% specificity in detecting these tumors.
    \item \textbf{Spleen tumors} \cite{valizadeh2024diagnostic}: A 2024 meta-analysis synthesized spleen tumor detection performance across different imaging modalities. On CT, radiologists achieved 77\% sensitivity and 91\% specificity in detecting spleen tumors.
    \item \textbf{Uterus tumors} \cite{franconeri2019asymptomatic}: In asymptomatic postmenopausal women on CT (22 cancers; 22 controls), the endometrium thickness was measured by radiologists to detect endometrial cancer. An 8 mm thickness threshold yielded 86\% sensitivity and 91\% specificity for detecting the tumors. This study considers only endometrial cancers, but our test dataset may include other types of uterus malignant tumors. 
    \item \textbf{Adrenal tumors} \cite{allard1990sensitivity}: The study considered 91 lung cancer patients. Of them, 53 had adrenal metastases (autopsy-validated). Radiologists could detect these metastasis on CT with sensitivity of 20 to 41\%, but high specificity (84 to 99\%). This study considers only metastasis, but our test dataset includes both metastasis and primary adrenal malignant tumors.
\end{itemize}




\end{appendices}


\bibliography{refs,zzhou}% common bib file

\end{document}